\documentclass[letterpaper, 10 pt, conference]{ieeeconf}  

\IEEEoverridecommandlockouts                              
\usepackage{cite}
\usepackage{graphics} 
\usepackage{epsfig} 
\usepackage{epstopdf}
\graphicspath {{./img/}}
\usepackage{mathptmx} 
\usepackage{times} 
\usepackage{amsmath} 
\usepackage{amssymb}  
\usepackage{gensymb}
\usepackage{multirow}
\usepackage{picinpar}
\usepackage{multicol}
\usepackage{stfloats}
\usepackage{caption}
\usepackage{graphicx}
\usepackage{subfigure}
\usepackage{wasysym}
\usepackage{textcomp}
\usepackage{float}
\usepackage{bm}

\usepackage[colorlinks,linkcolor=red,anchorcolor=blue,citecolor=green]{hyperref}

\usepackage{color}

\usepackage{array}
\usepackage{mdwmath}
\usepackage{mdwtab}
\usepackage{eqparbox}

\usepackage{xcolor,soul,framed} 
\usepackage{url}
\usepackage{subfigure}
\usepackage{diagbox}
\usepackage{booktabs}
\usepackage{amsfonts}
\usepackage{etoolbox}
\usepackage{makecell}

\makeatletter
\patchcmd{\@makecaption}
  {\scshape}
  {}
  {}
  {}
\makeatother
\def\tablename{Table}

\usepackage{algorithm}
\usepackage{algorithmic}
\usepackage{caption}

\newlength\myindent
\begin{document}
	
	\title{SE-Harris and eSUSAN: Asynchronous Event-Based Corner Detection Using Megapixel Resolution CeleX-V Camera}
	
	\author{Jinjian~Li$^{1}$, Chuandong Guo$^{1}$, Li~Su$^{2,*}$, Xiangyu Wang$^{2}$, Quan Hu$^{1,*}$
		\thanks{$^{1}$School of Aerospace Engineering, Beijing Institute of Technology, Beijing 100081, China}
		\thanks{$^{2}$Information Engineering College, Capital Normal University, Beijing 100048, China}
		\thanks{$^{*}$Corresponding~author:~li.su@cnu.edu.cn, huquan@bit.edu.cn}
	}

	
	
	
	\maketitle

\begin{abstract}

Event cameras are novel neuromorphic vision sensors with ultrahigh temporal resolution and low latency, both in the order of microseconds. Instead of image frames, event cameras generate an asynchronous event stream of per-pixel intensity changes with precise timestamps. The resulting sparse data structure impedes applying many conventional computer vision techniques to event streams, and specific algorithms should be designed to leverage the information provided by event cameras. We propose a corner detection algorithm, eSUSAN, inspired by the conventional SUSAN (smallest univalue segment assimilating nucleus) algorithm for corner detection. The proposed eSUSAN extracts the univalue segment assimilating nucleus from the circle kernel based on the similarity across timestamps and distinguishes corner events by the number of pixels in the nucleus area. Moreover, eSUSAN is fast enough to be applied to CeleX-V, the event camera with the highest resolution available. Based on eSUSAN, we also propose the SE-Harris corner detector, which uses adaptive normalization based on exponential decay to quickly construct a local surface of active events and the event-based Harris detector to refine the corners identified by eSUSAN. We evaluated the proposed algorithms on a public dataset and CeleX-V data. Both eSUSAN and SE-Harris exhibit higher real-time performance than existing algorithms while maintaining high accuracy and tracking performance.

\end{abstract}


\section{Introduction}
\label{sec:introduction}

Event cameras are bioinspired vision sensors that can capture scene motion in real time. Many previously difficult computer vision tasks, such as detecting and tracking objects in motion-blur or low-contrast scenes, can be appropriately solved by the asynchronous low-latency characteristics of event cameras. Unlike traditional frame-based cameras, event cameras only respond to changes in pixel brightness and output an asynchronous event stream at a high time resolution in the order of microseconds, greatly reducing the amount of redundant static data. Owing to their differences from frame-based cameras, event cameras have recently been applied in computer vision and robotics\cite{Muthusamy2020,Muthusamy2020b,mueggler2014event}. For example, an eye-in-hand grasping scheme for a robot manipulator based on an event camera has been proposed using a DAVIS240C event camera to support the control of a UR10 robot\cite{Muthusamy2020}, demonstrating the great potential of event cameras for robotic vision.

\begin{figure} [t]
	\centering 
	\subfigure[Event stream]{ 
		\includegraphics[width=1.55in]{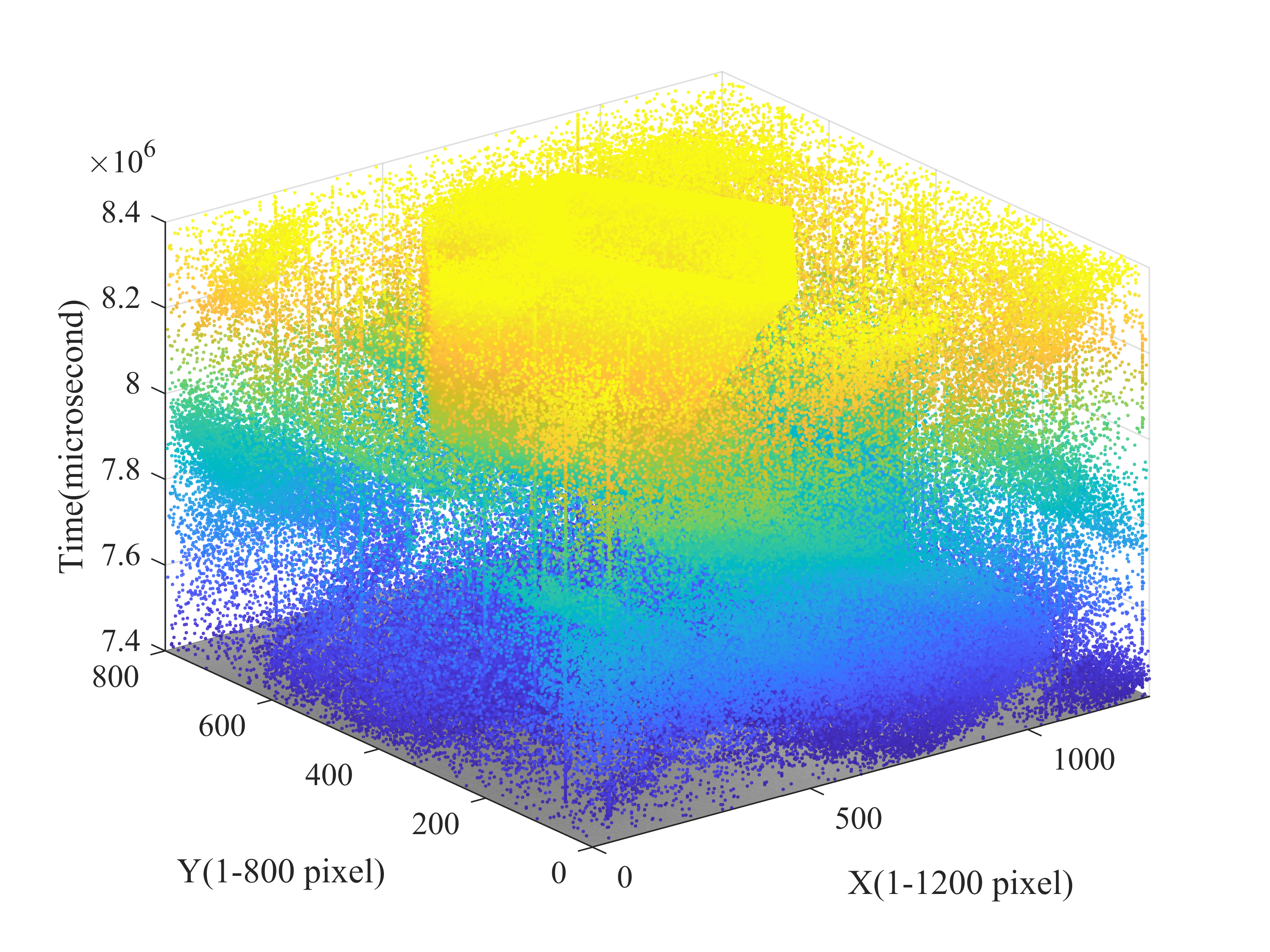} 
		\label{img:event_origin}
	}
	\subfigure[Corner stream]{ 
		\includegraphics[width=1.55in]{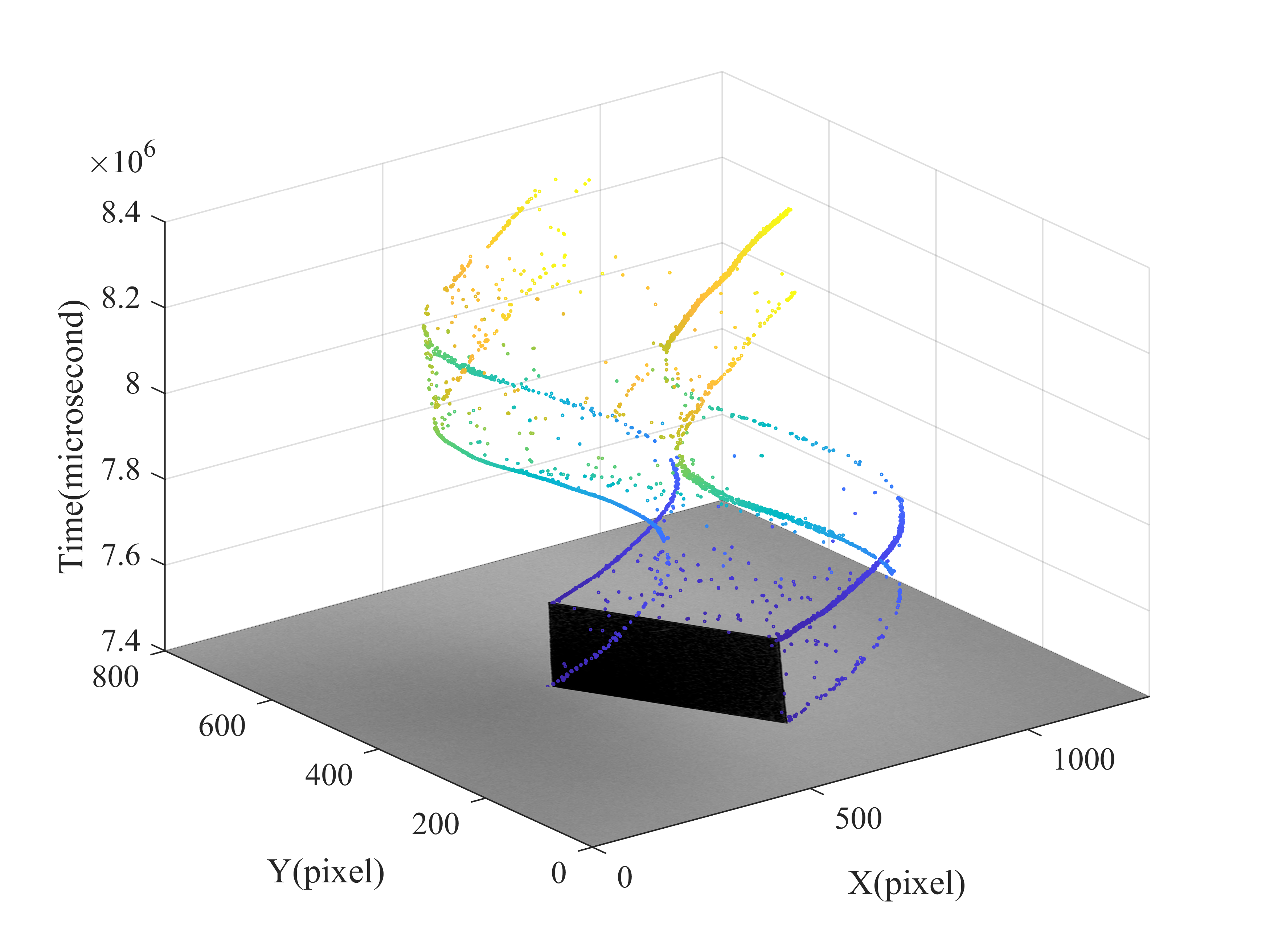} 
		\label{img:SE_Harris}
	}
	\captionsetup{font={small}}
	\caption{
		Event stream generated by high-resolution event camera. (a) Massive event data generated by CeleX-V event camera with megapixel resolution. (b) Results of SE-Harris detector, which combines eSUSAN and eHarris based on AED-SAE to extract corner events from massive event data. Note that the frame in the figure is just a supplementary description of the scene, and the proposed method does not use any frame information.
	}
	\label{img:Massive events}
\end{figure}

With the increasing commercialization of event cameras, various types have been released to the market, including DVS\cite{Lichtsteiner2008}, ATIS\cite{Posch2011}, and DAVIS\cite{Yang2014,Brandli2014}. Each pixel of DVS and ATIS processes visual information independently, while DAVIS combines asynchronous event streams with synchronous image frames. Other types of event cameras have recently been introduced, such as the Color-DAVIS346 color event camera\cite{moeys2017color} and the CeleX-VI/V optical flow event cameras\cite{huang2017dynamic,Huang2018}. CeleX-V provides a megapixel resolution of 1280 × 800 pixels, whereas the typical resolutions of DAVIS and ATIS are 346 × 260 and 480 × 360 pixels, respectively. High-resolution event cameras can respond to objects of various sizes and textures in complex scenes. Thus, they may allow to expand the application range and handle difficult computer vision tasks. However, high-resolution event cameras generate massive data, as shown in Figure \ref{img:Massive events}, thus imposing challenges to the efficiency of the corresponding algorithms. Moreover, a high-resolution event camera generates a great deal of noise events. Consequently, efficient and accurate filtering or denoising must be devised.

We consider real-time corner detection as an application of high-resolution event cameras such as CeleX-V. Various event-based asynchronous corner detectors are available, but most of them cannot handle the numerous events generated by high-resolution event cameras. For instance, event cameras can easily generate over 10 million events for a single scene. Considering the limitations of existing algorithms, we present a new detector for event data, eSUSAN, based on the conventional SUSAN (smallest univalue segment assimilating nucleus) algorithm. An adaptive time threshold extracts the univalue segment assimilating nucleus (USAN) area around each event. In addition, eSUSAN distinguishes corner events in the raw event stream through partitioning and pixel accumulation in the USAN area. Moreover, eSUSAN can quickly detect corner candidates without calculating gradients iteratively. Furthermore, adaptive normalization based on exponential decay is designed for the local surface of active events (SAE). This normalized local SAE is called AED-SAE and is constructed using a lookup table related to the adaptive time threshold and local SAE contrast. The AED-SAE is directly applied to an event-based Harris detector instead of the original binarized SAE, greatly reducing the computation cost for corner detection. As a result, we obtain a hybrid method called SE-Harris combining eSUSAN with an enhanced event-based Harris (eHarris) detector. The proposed method determines candidate corners using eSUSAN and refines them with a Harris detector based on AED-SAE. 

We evaluated the proposed method on public datasets to verify its effectiveness and demonstrate its remarkable advantages in high-speed and simple scenes. Then, we used the CeleX-V event camera to record megapixel event streams. Using the obtained dataset, the designability of the eSUSAN algorithm can be illustrated, and the real-time performance of eSUSAN, which exceeds that of existing algorithms, is demonstrated.

The remainder of this paper is organized as follows. Section \ref{sec:relatedwork} reviews related work on corner detection and SAE normalization. Section \ref{sec:method} describes the principle of the eSUSAN algorithm and SE-Harris detector. The experimental setup and performance results of the proposed algorithms are presented in Sections \ref{sec:Measures} and \ref{sec:Experimental}, respectively. Finally, we present conclusions and directions of future work in Section \ref{sec:conclusions}.

\section{RELATED WORK}
\label{sec:relatedwork}
In this section, we review related work on event-based corner detection, SAE normalization, and high-resolution event camera denoising.

\subsection{Event-Based Corner Detection} 

As event cameras provide a different data format from the conventional frame-based cameras, new algorithms should be devised for applications such as corner detection in an event stream. Event-based corner detection can be either direct or indirect\cite{Gallego2020,Mohamed2020}. 

Indirect detection relies on conventional computer vision algorithms applied to artificially synthesized frames by accumulating events\cite{zhu2017event,zihao2017event}. In this type of detection, event frames are generated over a period, and the Harris detector extracts corners\cite{harris1988combined}. Although indirect detection allows applying conventional algorithms to event camera data, improper time interval selection may occur. In fact, a fixed time interval is unsuitable for varying scene speed and can cause motion blur on images for fast objects. Furthermore, indirect detection may need the calculation of time-consuming optical flow and event lifetime. 

For improved corner detection and tracking, another type of indirect detection relies on dynamic and active pixel vision sensors\cite{tedaldi2016feature,Gehrig2018,Gehrig2020}. For this type of detection, frame information is used to detect initial features, and the event stream is used for feature matching and tracking. Although this type of detection is robust and achieves high tracking performance, motion blur or high-dynamic-range data may distort frame information. Moreover, indirect detection relying on frame information is deviated from the event-based processing paradigm and may hinder a timely response to new events, possibly causing delays.

To overcome the problems of indirect detection and tracking, various direct asynchronous detection algorithms have been proposed, including those based on gradients\cite{Vasco2016}, templates\cite{mueggler2017fast}, and learning\cite{manderscheid2019speed}. An early algorithm uses plane fitting to detect corner points and event characteristics\cite{Clady2015}. Most subsequent detection algorithms are based on the SAE\cite{Benosman2014a} (also called the time surface), which can reflect the spatiotemporal correlations in an event neighborhood. For instance, the Harris operator has been used to detect corners on the SAE\cite{Vasco2016}, which is binarized by the newest $N$ events. This method established the first gradient-based corner detector considering events. Although this method is adaptive to scene speeds, the resolution of event cameras or different scene textures impose the need for experiments to determine the value of $N$. In a subsequent development\cite{mueggler2017fast}, this method was improved by reserving ${N_l}$ events per pixel position instead of maintaining a fixed number of events. In \cite{Li2019,Mohamed2020a}, the global SAE is sorted to generate a local binary SAE and improve the computing efficiency. In addition, TLF-Harris\cite{Mohamed2020} simplifies the Harris score function into a product, further reducing the computation time. Overall, the gradient-based method is robust to noise but has a poor real-time performance.

The most representative template-based method for corner detection is eFAST\cite{mueggler2017fast}. It detects segments of contiguous pixels (arcs) in the event neighborhood using only pixel-level comparisons, improving the event processing speed. Subsequently, Arc*\cite{Ignacio2018} overcame the failure of eFAST to detect corners with wide angles. Arc* reduces the number of iterations per pixel but can be easily affected by noise because of its undemanding judgment. On the other hand, the parameters of these two methods are fixed, undermining their adaptation to different corner detection tasks.

To combine the advantages of different algorithms, the FA-Harris\cite{Li2019} and TLF-Harris algorithms have been proposed. They fully use a template-based algorithm to quickly filter edge events and a gradient-based algorithm to accurately locate the corner position. Alternatively, corner detection can be stated as a classification problem to naturally use learning-based methods, such as SILC\cite{manderscheid2019speed}, which classifies event streams through random forests. The detection results are continuously smooth and can be used for real-time corner tracking. Unfortunately, SILC achieves a detection rate of 1.6 Mev/s (megaevents per second), being insufficient to handle the data generated by high-resolution event cameras in real time.

\subsection{SAE Normalization} 
\label{subsec:normalization}

\begin{figure*} [t]
	\centering
	\subfigure{ 
		\includegraphics[width=6.0in]{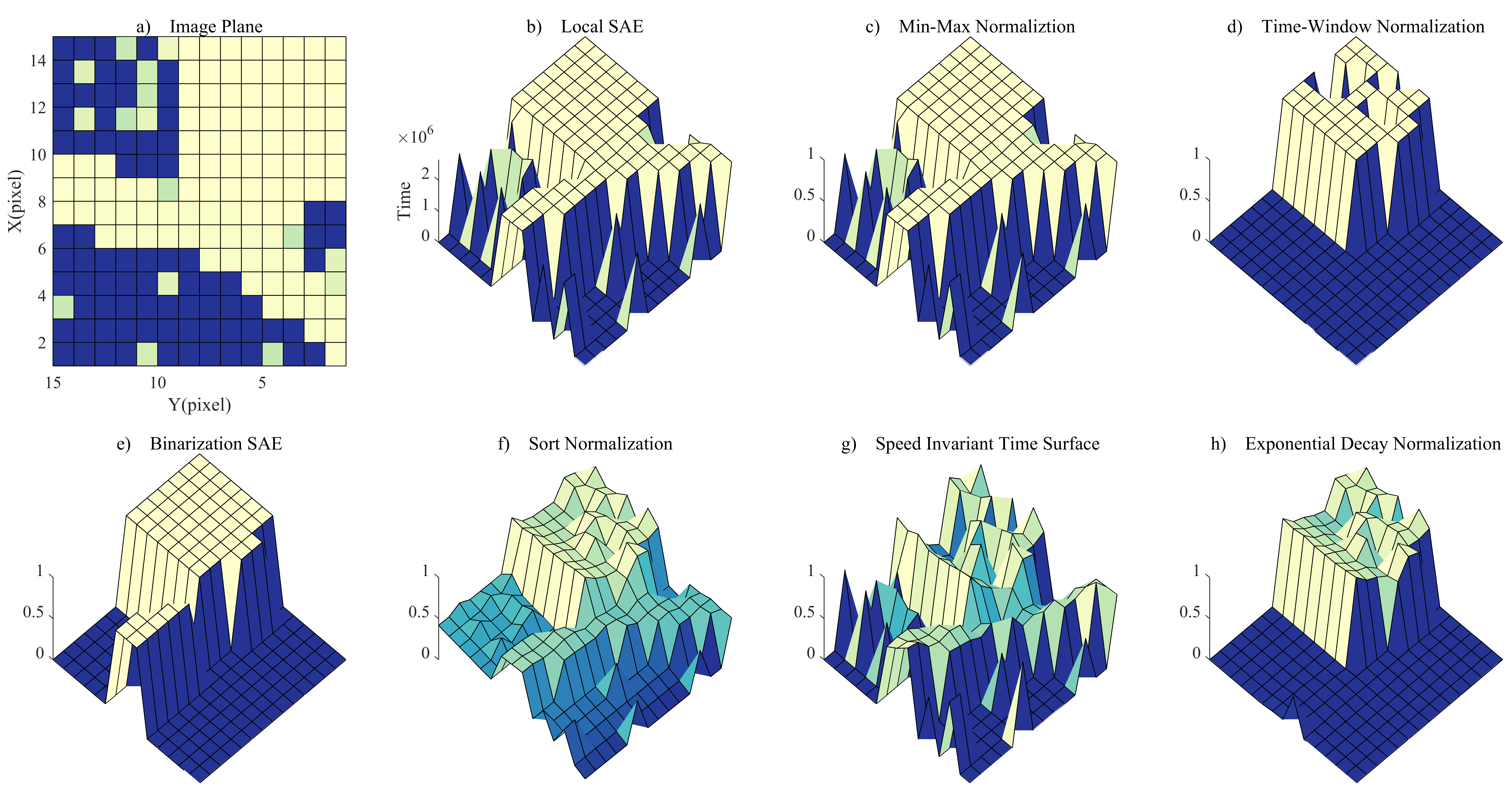} 
	} 
	\captionsetup{font={small}}
	\caption{SAE obtained from different normalization methods. (a) Local SAE with size of 15 × 15 projected on a two-dimensional plane. (b) Three-dimensional image of SAE. (c)–(h) SAEs generated by different normalization methods. In terms of noise-robustness and speed adaptability, exponential decay normalization shows a higher performance. Brighter colors indicate newer timestamps}
	\label{img:SAE}
\end{figure*}

Event cameras generate an asynchronous event stream, and each upcoming event can be described by equation (1), which describes its position, timestamp, and polarity:
\begin{equation}\label{event}
	{e_i} = \left[ {{x_i},{y_i},{t_i},{p_i}{]^T} = } \right[{{\bf{x}}_i},{t_i},{p_i}{]^T},~i \in \mathbb{N}
\end{equation}
where $i$ is the event index, ${{\bf{x}}_i}{\rm{ = }}[{x_i},{y_i}]$ represents the event position at absolute timestamp ${t_i}$, and ${p_i} \in \left\{ { - 1, + 1} \right\}$ is the polarity representing is the sign of the brightness change. As a single event does not carry enough information, asynchronous corner detection in an event stream requires the SAE, which reflects the global spatiotemporal event representation. The SAE can be obtained using mapping function  ${M_e}$ that maps each pixel location to its corresponding timestamps as follows:
\begin{equation}\label{con:sae}
	\begin{array}{*{20}{c}}
		{{M_e}:{^2} \to [0,\infty )},\\
		{{\bf{x}}:t = {M_e}({\bf{x}}).}
	\end{array}
\end{equation}
Most asynchronous corner detection methods are based on the SAE. However, for feature extraction, the monotonically increasing timestamps lead to classification failure when directly using the SAE. Therefore, efficient and accurate SAE normalization is required. We define a local SAE of the event ${e_i}$ as
\begin{equation}\label{con:localsae}
	{{\mathcal{A}}_i}({\bf{u}},p),
\end{equation}
where ${\bf{u}} = {[{u_x},{u_y}]^T}$ is the pixel distance to the current event, and $p$ is the SAE polarity of the SAE. Generally, the value range of $ux$ and $uy$ is take values in ${\rm{\{  - R, }}...{\rm{,R\} }}$, such that the size of the local SAE is ${\rm{(2R + 1)}} \times {\rm{(2R + 1)}}$. 

In Figure \ref{img:SAE}, we compare existing normalization methods. Some methods use timestamp contrast to directly normalize the SAE by applying min-max normalization\cite{Alzugaray2018}, time-window normalization\cite{afshar2019investigation}, and exponential decay normalization\cite{Lagorce2017}. From these methods, min-max normalization is widely used. However, for event cameras, the timestamp ranges from 0 to infinity, and min-max normalization cannot effectively reflect the timestamp contrast, especially when the local SAE has a large time span, as shown in Figure \ref{img:SAE}(c). Time-window normalization (also called event binning normalization) is specifically intended for event data. It divides the local SAE into two parts according to time window $\tau$, which requires empirical tuning. The value within time difference $\tau$ from the center pixel timestamp is regarded as 1, and the other values are regarded as 0. This process is described by equation \ref{con:timewin}. To restore the true shape of the time surface, this method can be improved by linear decay, as expressed in equation \ref{con:timewinlinear}.
\begin{equation}\label{con:timewin}
	{{\mathcal B}_i}({\bf{u}},p) = \left\{ {\begin{array}{*{20}{l}}
			{1,~{t_i} - {{\mathcal A}_i}({\bf{u}},p) \le \tau }\\
			{0,~{\rm{otherwise}}}
	\end{array}} \right.
\end{equation}

\begin{equation}\label{con:timewinlinear}
	\mathcal{L}_{i}(\mathbf{u}, p)=\left\{\begin{array}{ll}1+\frac{\mathcal{A}_{i}(\mathbf{u}, p)-t_{i}}{\tau}, & t_{i}-\mathcal{A}_{i}(\mathbf{u}, p) \leq \tau \\0, & \text { otherwise }\end{array}\right.
\end{equation}

However, time-window normalization has a poor adaptability to high-speed scenes, as an inappropriate value of $\tau$ generates an entire local SAE mapped to 1. On the other hand, not all similar times are mapped to 1, such as the situation illustrated in Figure \ref{img:SAE}(d), thus decreasing the detection accuracy. To avoid difficult parameter adjustment, exponential decay normalization has been proposed. It is similar to the time-window normalization and can be described as follows:
\begin{equation}
	{{\mathcal E}_i}({\bf{u}},p) = {e^{ - {\tau _e}({t_i} - {{\cal A}_i}({\bf{u}},p))}},
\end{equation}
Although attenuation factor ${\tau_e}$ is still related to the scene speed and should be determined experimentally, it is a simple method and can suppress noise.

More convenient normalization directly constructs speed-independent SAEs, with methods such as binary SAE\cite{Li2019}, sorting normalization\cite{Alzugaray2018}, and speed invariant time surface (SITS)\cite{manderscheid2019speed}. Binary SAE and sorting normalization are similar and require sorting out the local SAE according to the timestamp on the surface, but binary SAE maps the first ${N_l}$ timestamps to 1, while sorting normalization maps each timestamp to its own sort order, as shown in Figures \ref{img:SAE}(e) and \ref{img:SAE}(f), respectively. In addition, binary SAE conveniently generates normalized SAEs from the global SAE, but parameter ${N_l}$ influences detection. Although these two normalization methods effectively circumvent speed information, the required sorting is time-consuming. 

Alternatively, SITS is a novel method for SAE normalization that has been successfully used for learning-based corner detection. The SITS algorithm proceeds as follows. Set window width R. When a new event occurs, if each pixel in the ${\rm{(2R + 1)}} \times {\rm{(2R + 1)}}$ neighborhood is larger than $S\left( {0,p} \right)$, subtract 1. Finally, set the position where the new event occurred to ${\left( {2R\; + \;1} \right)^2}$. This method is efficient but highly sensitive to the value of $R$. When $R$ is small, SITS is susceptible to noise. When $R$ is large, the real-time performance declines. As shown in Figure \ref{img:SAE}(g), for ${\rm{R  =  4}}$, noise exists for a long period and interferes with SAE normalization. Especially for the CeleX-V event camera, the influence of noise is further amplified.

Other recent normalization methods have been proposed, such as chain SAE\cite{Lin2020} and ATSLTD\cite{Chen}. Although they are both driven asynchronously, the former is more suitable for on-demand tasks, while the latter is used to generate high-contrast artificial frames.

\subsection{Denoising Event Camera Data}

Early denoising an event stream can substantially improve the real-time performance and accuracy of visual recognition algorithms. For high-resolution event cameras, which often produce noisy data, efficient denoising can considerably improve corner detection. Noise is generally caused by thermal influences and junction leakage currents\cite{Lichtsteiner2008} also called background activity. Unlike real events, background activity does not have temporal or spatial correlations, and can thus be removed. 

The most efficient denoising algorithms are the nearest-neighbor-based filters in \cite{Delbruck2008} and \cite{liu2015design}. The algorithm in \cite{Delbruck2008} removes events that have no events in the surrounding eight pixels in the previous fixed-time window. The algorithm in \cite{liu2015design} divides the window size of ${{\rm{s}}^2}$ pixels into a group and stores the most recent timestamp of the group. This algorithm can reduce memory consumption and running time. In addition to using the nearest neighbors, some algorithms such as those proposed in \cite{Feng2020,Wu2020} use large spatiotemporal windows to finely filter background activity. The algorithm in \cite{Feng2020} uses event density, and that in \cite{Wu2020} builds a probability undirected graph to represent denoising as an optimization problem. These two algorithms provide accurate denoising, but their computational overheads are large. 

Aiming at the multiple trigger characteristics of real events, eFilter has been proposed\cite{Ignacio2018}. It smooths the time surface by filtering repeated events within a period. Unlike eFilter, the GF adaptive denoising algorithm\cite{Guo2020} uses global spatiotemporal information to calculate an adaptive time threshold and treats events larger than this threshold in a ${{\rm{s}}^2}$ window as noise. The adaptive time threshold is expressed as:
\begin{equation}\label{con:tgf}
	TGF = \frac{{TD \times (X \times Y)}}{{{s^2} \times FN \times SF}},
\end{equation}
where $TD$ is a time interval, $X$ and $Y$ are the length and width of the event camera images, respectively, $s$ is the size of the subsampling window, $FN$ is the number of events that occur in time interval $TD$, and $SF$ is a scaling factor used to distinguish the time interval of noise events. Parameter $SF$ counts the time interval between adjacent events and can distinguish them from noise. As global information is not updated, real-time performance can be achieved. Therefore, we adopt a GF filter to preprocess the event stream provided by the CeleX-V camera. The lack of polarity information causes the corners at the polarity junction to be incorrectly filtered when using eFilter. Thus, we use adaptive time threshold $TGF$ for corner detection. Some problems in the calculation of $TGF$ are explained in Section \ref{sec:method}.

\begin{figure} [t]
	\centering 
	\subfigure[]{ 
		\includegraphics[width=1.55in]{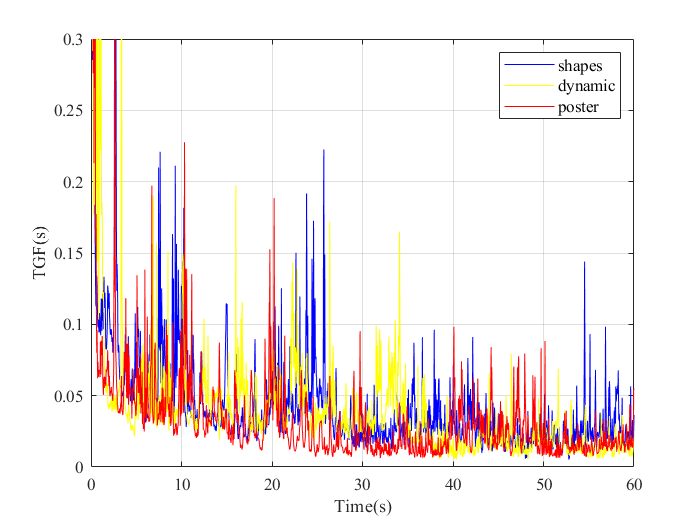} 
		\label{img:TGF_time}
	}
	\subfigure[]{ 
		\includegraphics[width=1.55in]{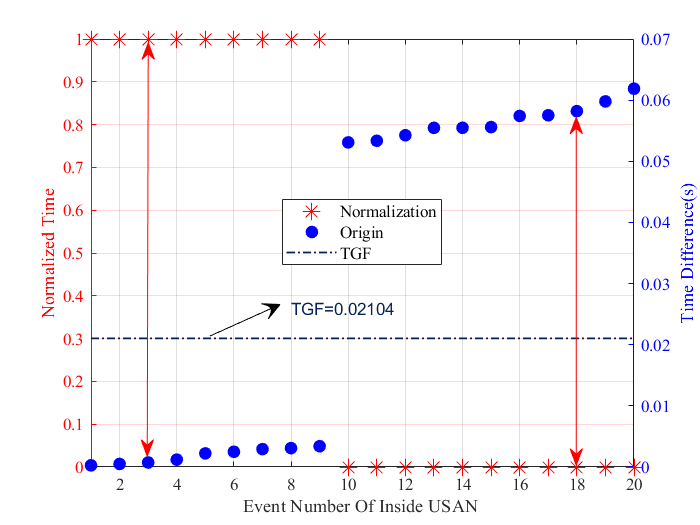} 
		\label{img:TGF_principle}
	}
	\captionsetup{font={small}}
	\caption{
		$TGF$ values over time according to scene speed. (a) $TGF$ variation in three different texture scenarios. (b) According to TGF, time-related events can be clearly distinguished. Blue points represent the timestamp between the inner-circle neighborhood and the center pixel (from small to large), and red points represent events normalized by the proposed normalization method. AED-SAE is constructed using the red points.
	}
	\label{img:TGFprinciple}
\end{figure}
\section{Proposed Event-Based Corner Detection}
\label{sec:method}

\subsection{Improved Adaptive Time Threshold}
\label{subsec:TGF}

The principle of the GF filter is described above. Apart from separating real events from noise, we expect that adaptive time threshold $TGF$ allows to represent event similarity, which is defined by close locations and similar timestamps. For timestamps, $TGF$ can adaptively separate events in scenes with objects moving at different speeds. Intuitively, when the relative speed between the scene and camera is high, the number of events increases and TGF decreases, and vice versa. 
As $TGF$ is also closely related to the scene texture, we can rewrite it as follows:
\begin{equation}\label{con:tgfpro}
	TGF({\bf{v}}) = \frac{{{T_{\rm{c}}}}}{{\lambda (s) \cdot {N_{\rm{e}}}(s,{\bf{v}})}}\;,
\end{equation}
where ${T_c}$ is the constant part of equation \ref{con:tgf}, and ${{N_{\rm{e}}}}$ is the total number of events generated in a period and expressed as a function of scene texture density and scene speed. We expect that $TGF$ is only related to the scene speed. Therefore, we should design a scaling factor $\lambda$ to offset the influence of scene texture. Specifically, $\lambda$ should be estimated by calculating the spatial density of an event. Considering real-time corner detection, we manually select the scaling factor for different texture scenes during evaluation. To overcome $TGF$ variations that affect corner detection, we smooth $TGF$ as follows:
\begin{equation}\label{con:tgfsmooth}
	TG{F_{\rm{j}}} = 0.05 \cdot TG{F_{{\rm{j - 1}}}} + 0.95 \cdot \frac{{{T_{\rm{c}}}}}{{\lambda  \cdot {N_{\rm{e}}}_{\rm{j}}}}.
\end{equation}

Figure \ref{img:TGFprinciple} shows variations in $TGF$ over time in scenes with three different textures. The texture levels ranging from high to low corresponding to scene subsets shapes, dynamic, and poster, for which we set $\lambda$ to 5, 2, and 1, respectively. As a results, $TGF$ maintains the same order of magnitude regardless of the texture.

\subsection{eSUSAN For Event Camera}
\label{subsec:esusan}

\begin{figure*} [t]
	\centering 
	\subfigure[]{ 
		\includegraphics[width=1.6in]{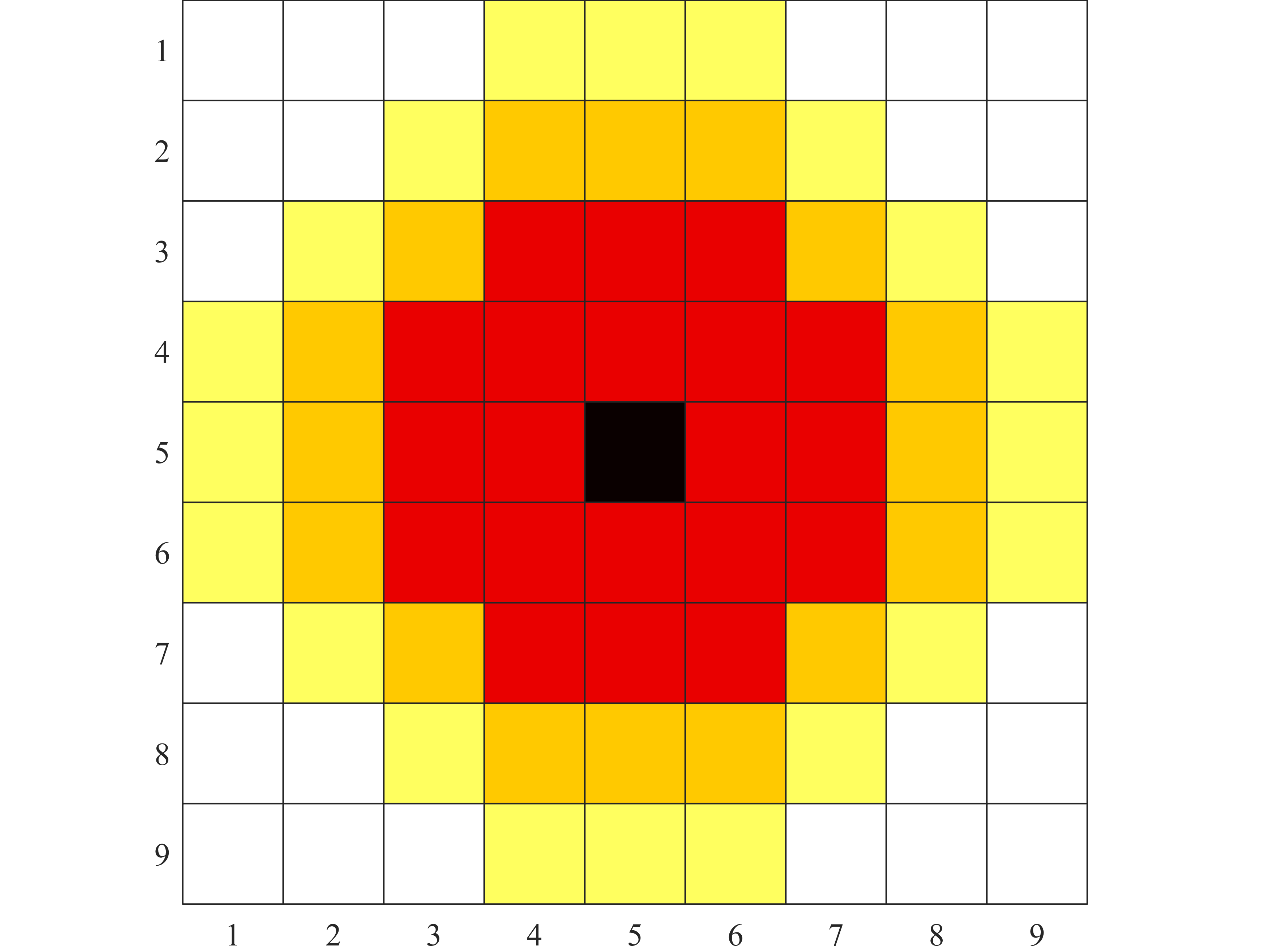} 
		\label{img:kernel}
	}
	\subfigure[]{ 
		\includegraphics[width=1.6in]{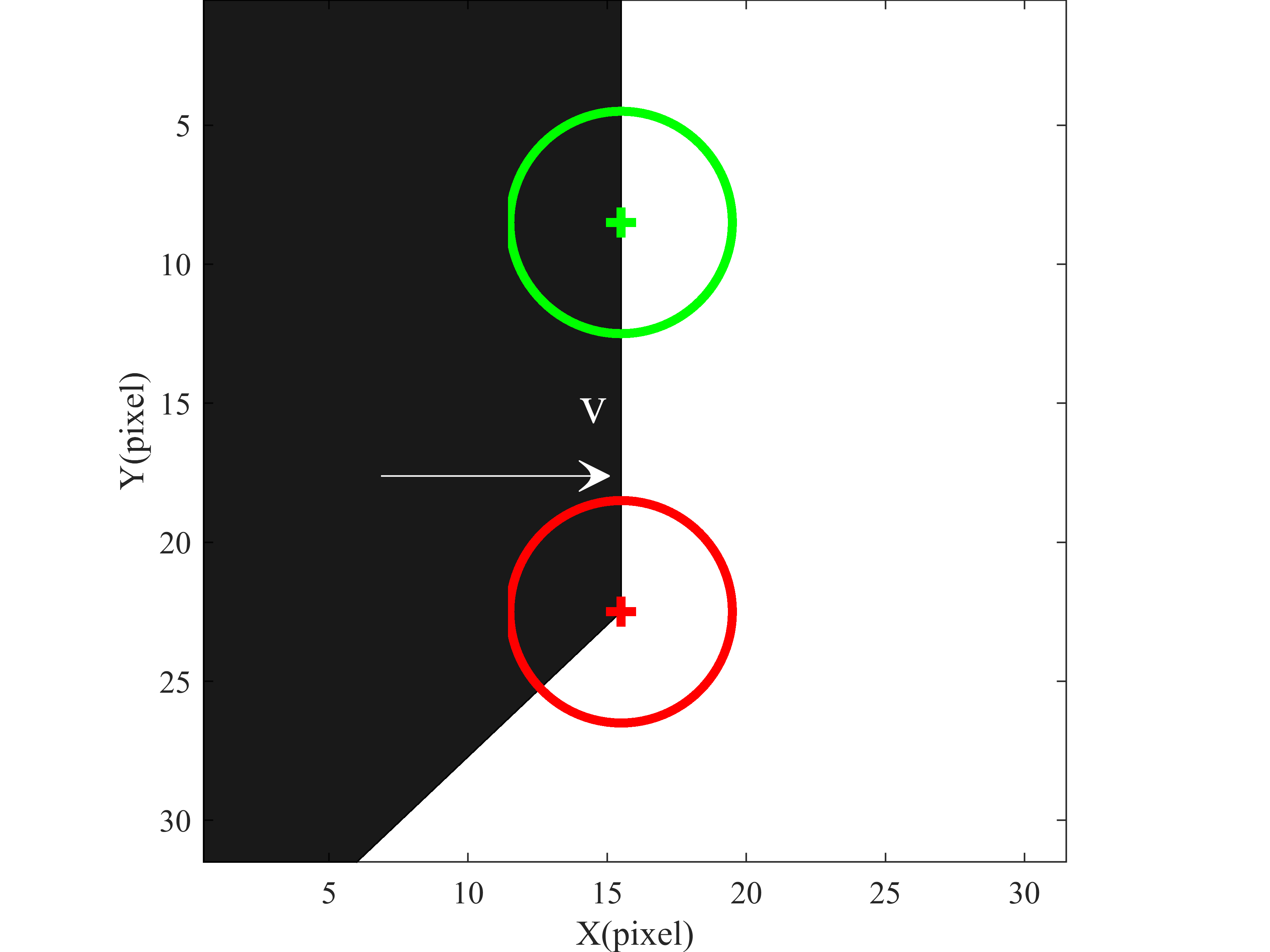} 
		\label{img:corner_move}
	}
	\subfigure[]{ 
		\includegraphics[width=1.6in]{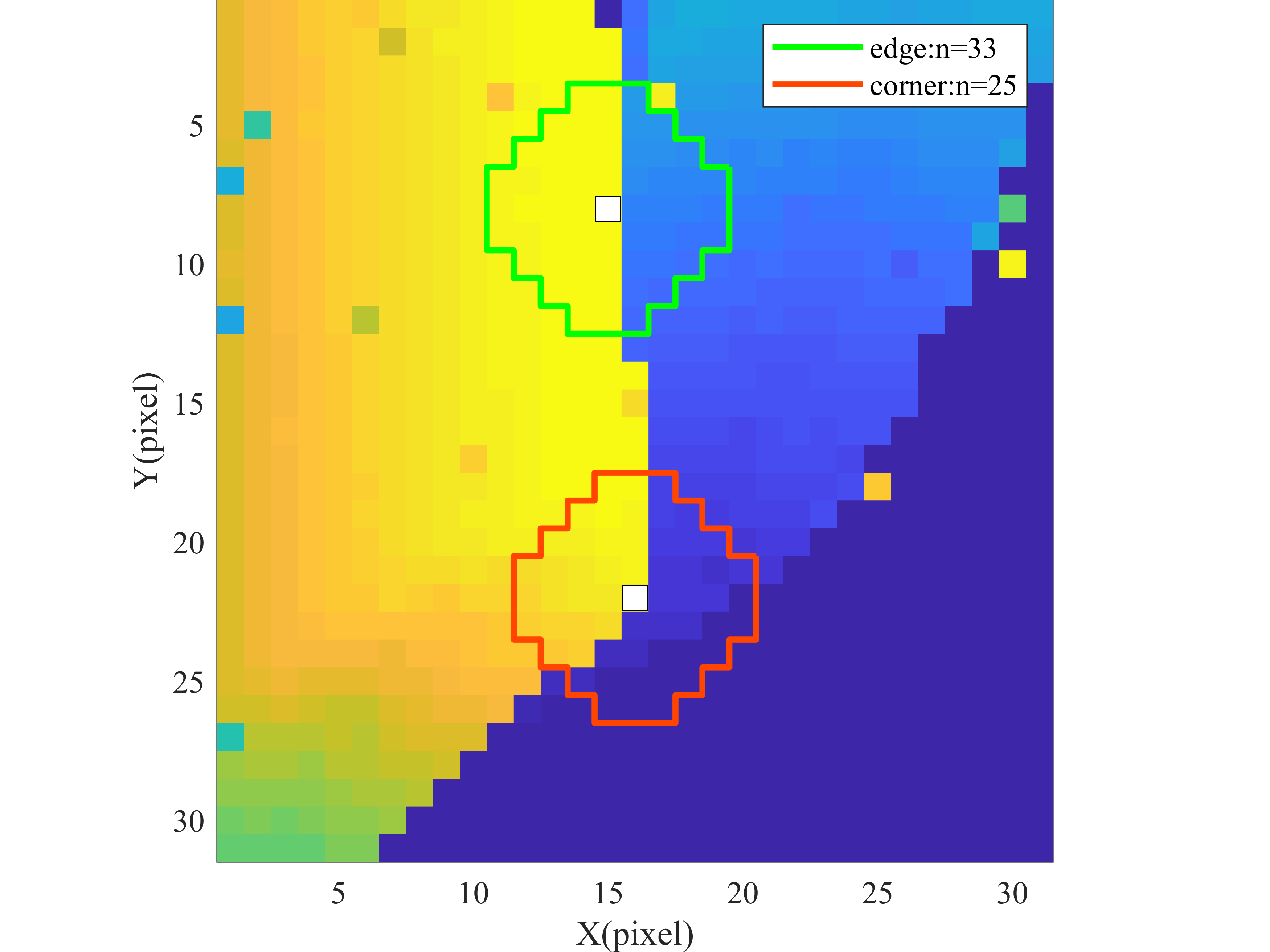} 
		\label{img:corner_sae_2d}
	}
	\subfigure[]{ 
		\includegraphics[width=1.6in]{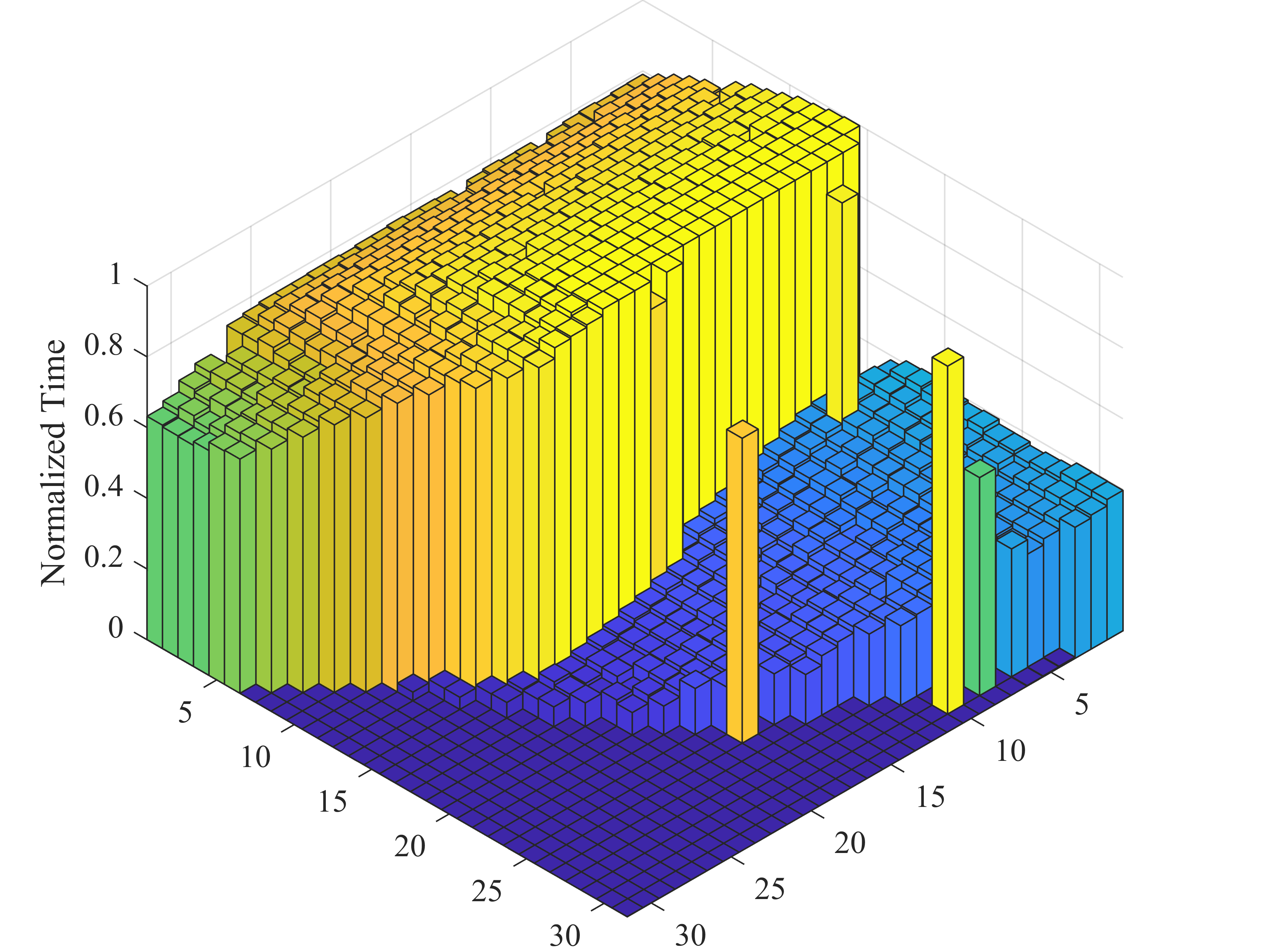} 
		\label{img:corner_sae_3d}
	}
	\captionsetup{font={small}}
	\caption{Principle of proposed eSUSAN for event-based corner detection. (a) A circle kernel with a radius of 4 and divided into three parts is used to inspect from inside to outside. (b) Edges and corners of the object move to the right and are identified using the circular kernel. (c) The SAE (31 × 31) is formed by the movement of edges and corners. Newer timestamps are closer to yellow, and older ones are closer to blue. In pixel coordinates, the circular core consists of 57 pixels. In addition, 33 similar pixels can be detected in the green core, and 25 similar pixels can be detected in the red core. The similar pixels can be used to distinguish edges and corners. The number of edges does not reach half of the template pixels. (d) Three-dimensional spatiotemporal SAE.
	}
	\label{img:eSUSANprinciple}
\end{figure*}

We mention above two template-based corner detection algorithms that only use arc information to detect corners. eFast is relatively strict but cannot detect a wide range of angles, while Arc* relaxes the judgment conditions but is more susceptible to noise. To reliably detect corners, we should use all the information of the circular template (kernel or window), including the area and center of gravity, except for the arc.

We improve the conventional SUSAN algorithm\cite{smith1997susan} for use with event camera data. The proposed eSUSAN algorithm counts the number of pixels that are similar to the center point in a circular template with a radius of 3 or 4 to distinguish the corners and edges, as illustrated in Figure \ref{img:eSUSAN}. The original SUSAN algorithm proved that a circular template with a radius of 3 is sufficient for corner detection. To be consistent with the previous event-based detector, we use a circle with a radius of 4 but slightly modify the detection process. 

As the event stream does not have grayscale information, we directly compare the similarity between timestamps to extract the USAN. There are many methods for determining the similarity threshold without calculating the optical flow like using SITS, and the similarity threshold is ${\rm{(2R  +  1) }} \times {\rm{R}}$ considering a constant surface after a slope. In our work, we uses $TGF$ as the similarity threshold directly because $TGF$ reflects the global optical flow. The USAN area can be simply determined as:
\begin{equation}\label{con:usan}
	c({{\bf{x}}_i},{\bf{u}}) = \left\{ {\begin{array}{*{20}{c}}
			{1,~{t_i} - {{\cal A}_i}({{\bf{x}}_i} + {\bf{u}},p) \le TGF}\\
			{0,~{t_i} - {{\cal A}_i}({{\bf{x}}_i} + {\bf{u}},p) > TGF}
	\end{array}} \right..
\end{equation}

After dividing the USAN area, we count the number of pixels, $n$, in the USAN area and distinguish the corner points according to $n$. The generated events only reflect the information of moving objects in an event camera. Thus, geometric threshold $g$ differs from than in the frame-based SUSAN algorithm. Except for noise, as most events are located on the edges and corners, the inner part of an object and the intra-pixel edge case mentioned in the original algorithm can be disregarded. Then, as the filter cannot completely remove noise, the low limit of $g$ should be determined. Finally, the USAN area at the turning edge cannot be filled, and the circular kernel should be modified. In the proposed method, the kernel is divided into three parts from inside to outside for detection. If one part does not satisfy the condition on $n$, it cannot be considered as a corner. We calculate $n$ as follows:
\begin{equation}\label{con:n}
	{n_r}({{\bf{x}}_i}) = \sum\limits_{{{\bf{u}}_r}} {c({{\bf{x}}_i},{\bf{u}})} {\rm{ ,  }}r = 1,2,3,
\end{equation}
where $r$ represents the different parts shown in Figure \ref{img:kernel}. This formulation can minimize false edge detection caused by a change in motion direction. As the original algorithm, we set geometric threshold $g$. Note that $n$ of the edges is not half of the pixel number in the circle kernel. To better filter the edge and retain most corner points, we set $\frac{1}{2}{n_{\max }}$ and ${n_{{\rm{egde}}}}$ as the upper and lower bounds of $g$, respectively. We also set ${g_{{\rm{noise}}}}$ to distinguish noise events, which are usually $\frac{1}{8}{n_{\max }}$. For the three parts, ${n_{\max }}$ is 21, 37, and 57, and ${n_{{\rm{egde}}}}$ is 13, 22, and 33, respectively. As in the original algorithm, we set geometric threshold $g$ heuristically to detect corners with different sharpness, rendering eSUSAN more adaptable than other template-based methods.

\subsection{AED-SAE for Harris Detector}
\label{subsec:AED}

The global SAE makes eHarris efficient but generates a local normalized SAE that demands time-consuming sorting operations. We propose AED-SAE to directly extract local patches from the global SAE without sorting. The proposed method is similar to exponential decay normalization but does not require manual adjustment of the attenuation coefficient. In addition, it can increase the timestamp contrast for the final time surface to be similar to the binarized time surface, thus promoting corner detection. This normalization method is also inspired by the USAN region-solving method proposed in \cite{smith1997susan}. We perform direct normalization as follows:
\begin{equation}\label{con:aed}
	{{\cal E}'_i}({\bf{u}},p) = {e^{ - {{(\frac{{{t_i} - {{\cal A}_i}({\bf{u}},p)}}{{\tau  \cdot TGF}})}^6}}},
\end{equation}
where ${{\cal A}_i}({\bf{u}},p)$ is the local SAE formed in area $(2R + 1) \times (2R + 1)$ around event $i$, $TGF$ is the time threshold, and $\tau$ is a parameter usually taken as 1. With $TGF$ in the exponent denominator, the resulting event surface adapts to different scene speeds, whereas the original method requires manual parameter adjustment for according to the speed. To use Harris corner detection on this surface, we use the sixth power to further smooth the normalization function for the final event surface to be close to the binarization result, as shown in Figure \ref{img:TGF_principle}. Moreover, smoothing can suppress the influence of parameter changes. As this formula is complicated and its calculation is time-consuming, we use the corresponding two-dimensional lookup table to construct the AED-SAE. 

Then, we directly use the Harris corner detection operator on the AED-SAE. For the latest event, ${e_i}$, symmetric Harris matrix $M$ can be calculated as follows:
\begin{equation}\label{con:Harrismatrix}
	M({e_i}) = \sum\limits_{{{{\cal E}'}_i}({\bf{u}},p)} {g({\bf{u}})\nabla I({\bf{u}})} {\nabla ^T}I({\bf{u}}),
\end{equation}
where ${\rm{g(u)}}$ is the Gaussian window function, and ${\nabla I({\bf{u}})}$ is the gradient of the local patch convolved with the Sobel kernel. Furthermore, Harris score $H$ is given by
\begin{equation}\label{con:Harrisscore}
	H = \det (M) - k \cdot Trace{(M)^2},
\end{equation}
where $k = 0.04$ is a parameter defined heuristically. If $H$ is larger than a threshold, the latest event ei is classified as a corner. We call AED-eHarris to this improved event-based Harris detector.

\subsection{Down-sampling SAE}
\label{subsec:downsampling}

\begin{figure} [t]
	\centering 
	\subfigure[]{ 
		\includegraphics[width=1.55in]{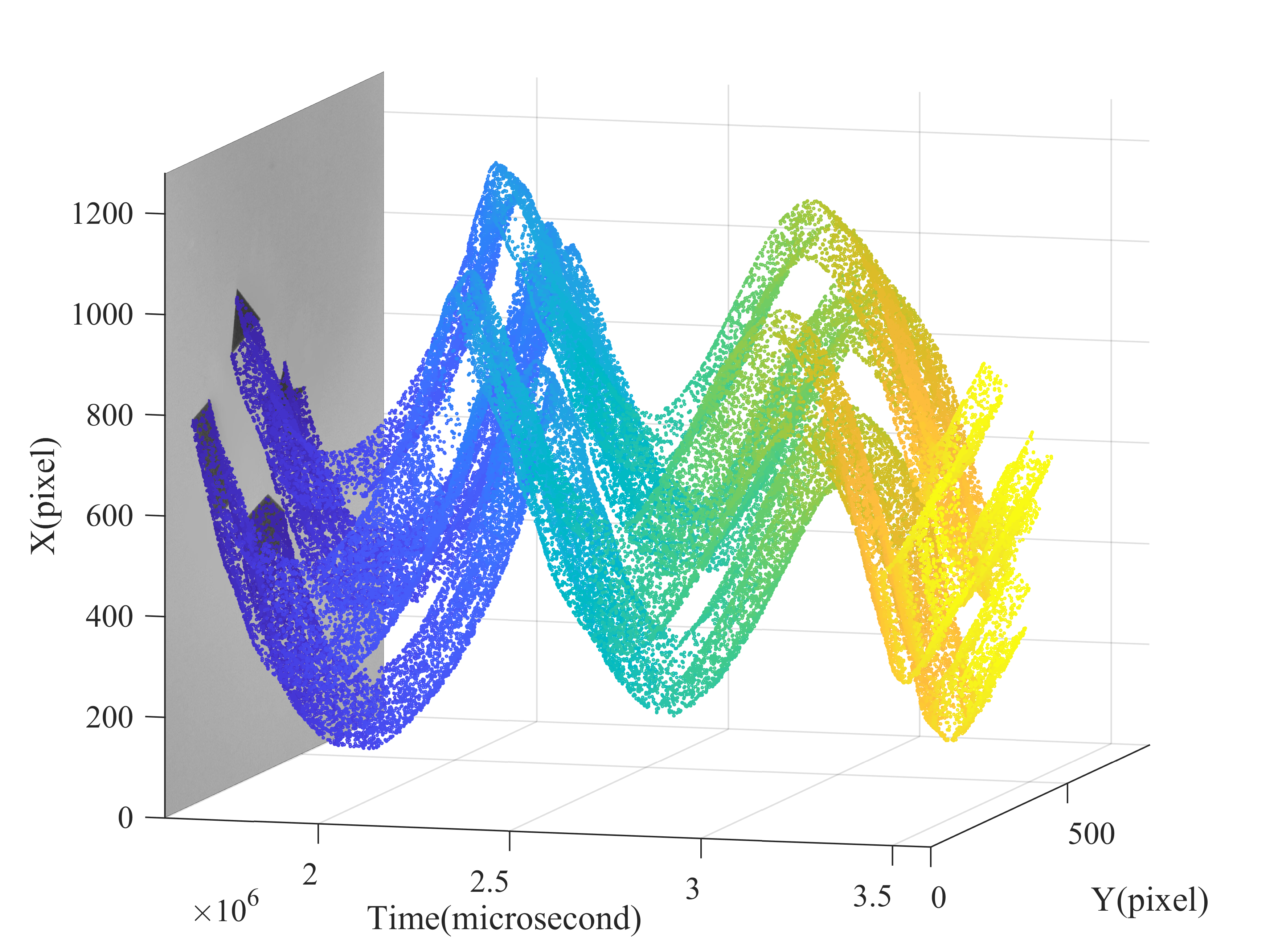} 
		\label{img:AED_no_s2}
	}
	\subfigure[]{ 
		\includegraphics[width=1.55in]{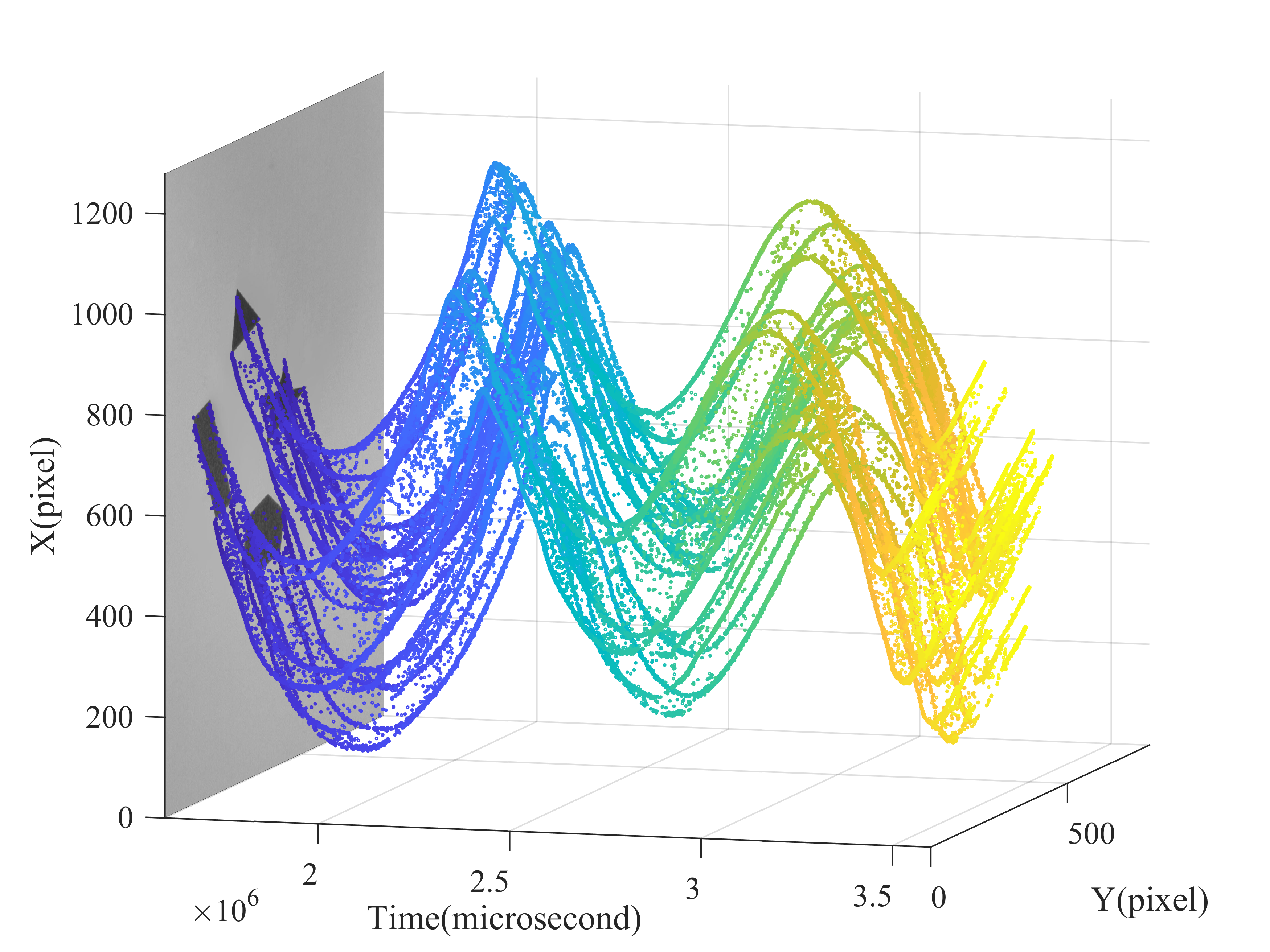} 
		\label{img:AED_s2}
	}
	\captionsetup{font={small}}
	\caption{
		The impact of SAE downsampling on corner detection. Use AED-eHarris to detect corner points in Celex-V dataset. (a) The detected corner event stream using the original SAE. Although the true corner can be detected, there are many cases of false detection of edges, and the detection performance is relatively low. (b) The detection result by using down-sampling SAE and the corner points are restored to the real pixel plane. At this time, most of the edges are filtered and the corners are very coherent.}
	\label{img:downsampling}
\end{figure}

Event-based corner detectors should apply to event cameras of any resolution. In other words, they should be scale-invariant, which is not mentioned in previous work. Previous detectors were only verified on low-resolution event datasets, but for high-resolution event cameras, their direct application will be problematic. When a high-resolution event camera shoots objects with the same proportion in the pixel plane, more events will be generated on the same edge, which often makes jagged noise on the edge identified by mistake as a corner, such as Figure \ref{img:AED_no_s2}. While high-resolution event cameras provide more precise visual information, we often focus only on the optimal corner points. To solve this problem, we take a down-sampling factor $s = 2$ to construct a layer of down-sampling SAE. Each incoming event is mapped to the corresponding down-sampling space and can be judged by TGF whether it is a real event. When real event signals appear, the down-sampling SAE is updated and used to corner detection subsequently. It should be noted that down-sampling reduces the resolution of SAE, which not only significantly improves the multi-scale performance but also reduces the calculation of corner detectors, making them more suitable for high-resolution tasks, as shown in Figure \ref{img:AED_s2}.

\section{Experimental Setup}
\label{sec:Experimental}

\subsection{Evaluation Datasets}
\label{subsec:Datasets}

We used two datasets to evaluate the proposed algorithm, namely, the public dataset in \cite{Mueggler2017} and a CeleX-V dataset that we recorded. The event camera dataset in \cite{Mueggler2017} was collected using DAVIS240C at a resolution of 240 × 180 pixels and includes asynchronous events, intensity images, and inertial measurements. For a fair comparison, we tested detectors in both simple and complex and low and high-texture scenes, including scene subsets \emph{shape}, \emph{dynamic}, and \emph{poster}.

\begin{figure} [t]
	\centering 
	
	\subfigure[Origin event]{ 
		\includegraphics[width=1in]{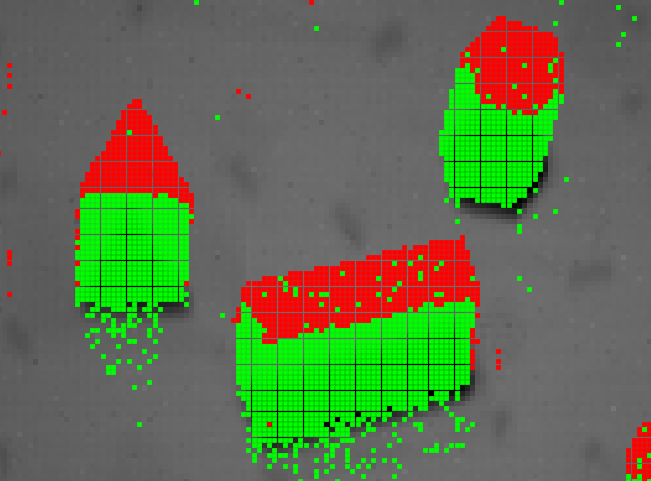} 
		\label{img:origin_shapes}
	}
	\subfigure[Geharris$^{*}$ with 5 × 5 sobel]{ 
		\includegraphics[width=1in]{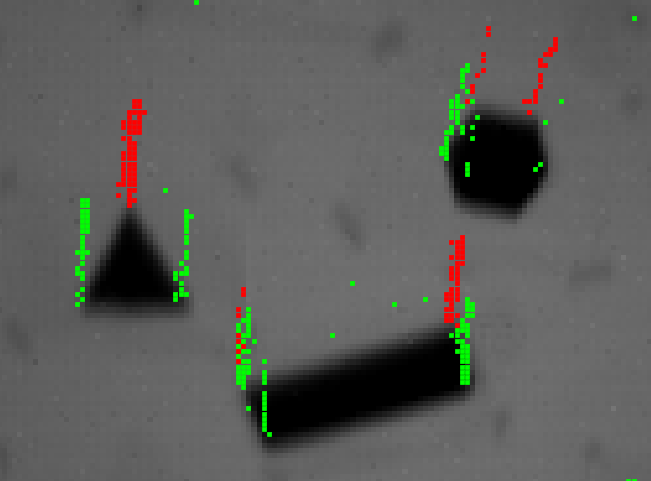} 
		\label{img:Geharris_shapes_5_8}
	}
	\subfigure[Geharris$^{*}$ with 7 × 7 sobel]{ 
		\includegraphics[width=1in]{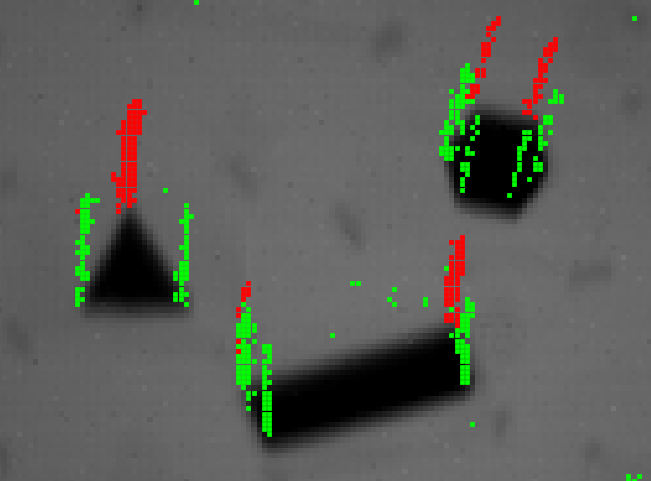} 
		\label{img:Geharris_shapes_7_16}
	}
	\subfigure[eFAST$^{*}$]{ 
		\includegraphics[width=1in]{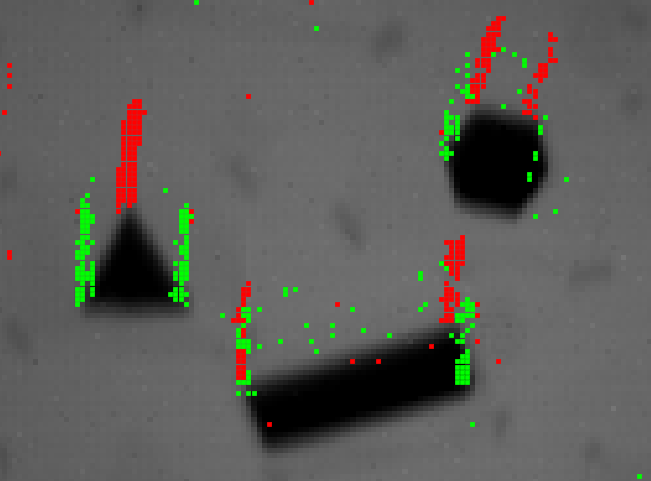} 
		\label{img:eFAST}
	}
	\subfigure[Arc$^{*}$]{ 
		\includegraphics[width=1in]{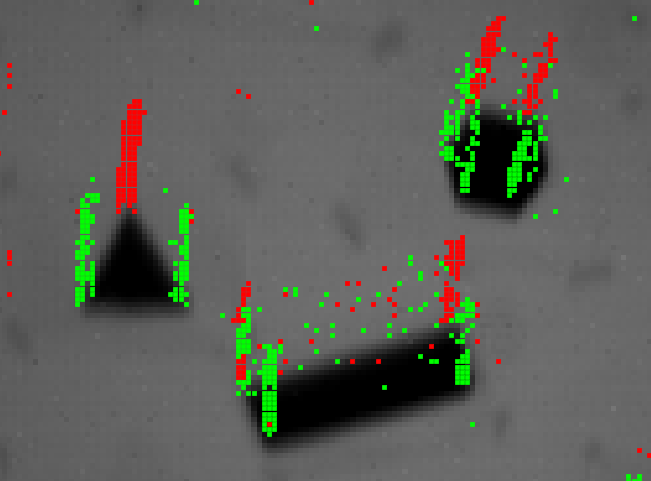} 
		\label{img:Arc}
	}
	\subfigure[eSUSAN$^{*}$]{ 
		\includegraphics[width=1in]{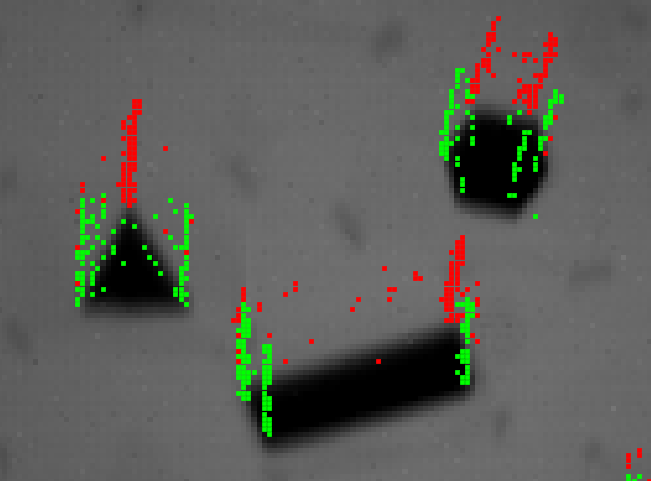} 
		\label{img:eSUSAN}
	}
	\captionsetup{font={small}}
	\caption{Accumulated 30 ms of original events and corner events detected by various methods using subset shapes. Red and green points represent positive and negative motion, respectively. (a) Original events. (b) Corner events detected by G-eHarris* with 5 × 5 Sobel kernel and threshold of 8. Several corners should be filtered in a large range. (c) Results using 7 × 7 Sobel kernel and threshold of 16 to improve detection. (d)–(f) Results of eFAST*, Arc*, and eSUSAN*, respectively. Unlike Arc* and eSUSAN*, eFAST* has a limited ability to detect all the corners. 		
	}
	\label{img:corners}
\end{figure}

\subsection{Evaluation Measures}
\label{sec:Measures}

There are various evaluation measures for event-based corner detectors. The average time spent to process a single event and the reduction rate can be used to evaluate the processing performance. If a detector provides both a higher event reduction rate and less computational overhead, it is more suitable as a corner event stream input for advanced algorithms. Note that a high event reduction rate does not necessarily indicate the suitability of a detector. 

The detection quality can be obtained from the false positive rate, true positive rate (TPR), accuracy, and 3D reprojection error. We used the same definition of accuracy available in \cite{Mohamed2020,Li2019}. Specifically, the accuracy of a corner detector is defined as the ratio of internal cylindrical corners to the total internal cylindrical corners formed by the true trajectory. It can be calculated as 
\[TP/\left( {TP{\rm{ }} + {\rm{ }}FP} \right),\]
where TP (true positives) is the number of corner events that fall on a small (inner) cylinder with a radius of 3.5 pixels, and FP (false positives) is the number of events between the small cylinder and a large (outer) cylinder with a radius of 5 pixels. Two oblique cylinders were constructed using the true trajectory generated by the Harris and Kanade–Lucas–Tomasi algorithms \cite{lucas1981iterative}. We also evaluated the TPR of our detectors as in \cite{Ylmaz2021}. The TPR is only suitable for datasets collected using the DAVIS camera, as they contain intensity information. For instance, the TPR is inappropriate for DVS, which cannot generate frames and events synchronously, or for high-speed and high-dynamic-range scenes.

We can only evaluate the continuity of the corner event stream in the dataset collected using the CeleX-V event camera with measures such as the 3D reprojection error, lifetime of the corner trajectory, and detection validity. These measures assume that a corner detector can provide continuous corner events without any complex trackers given the high time resolution of the event camera. As the 3D reprojection error uses the camera pose information to estimate the homography between two different timestamps, it can more accurately validate the tracking trajectory. We only used the lifetime of the corner trajectory and detection validity for evaluation. As in \cite{manderscheid2019speed,Ylmaz2021}, we used nearest-neighbor matching to obtain the tracking result with $r$, $\delta t$, and $n$ set to 3 pixels, 5 $ms$, and 10 events, respectively. We considered that a length above 10 to be a valid trajectory, and the events on the trajectory were valid corner events.

\section{Experimental Results}
\label{subsec:Results}

\begin{figure} [t]
	\centering 
	\subfigure[Event stream]{ 
		\includegraphics[width=1.55in]{./img/origin.png} 
		\label{img:celex_origin}
	}
	\subfigure[GF filter]{ 
		\includegraphics[width=1.55in]{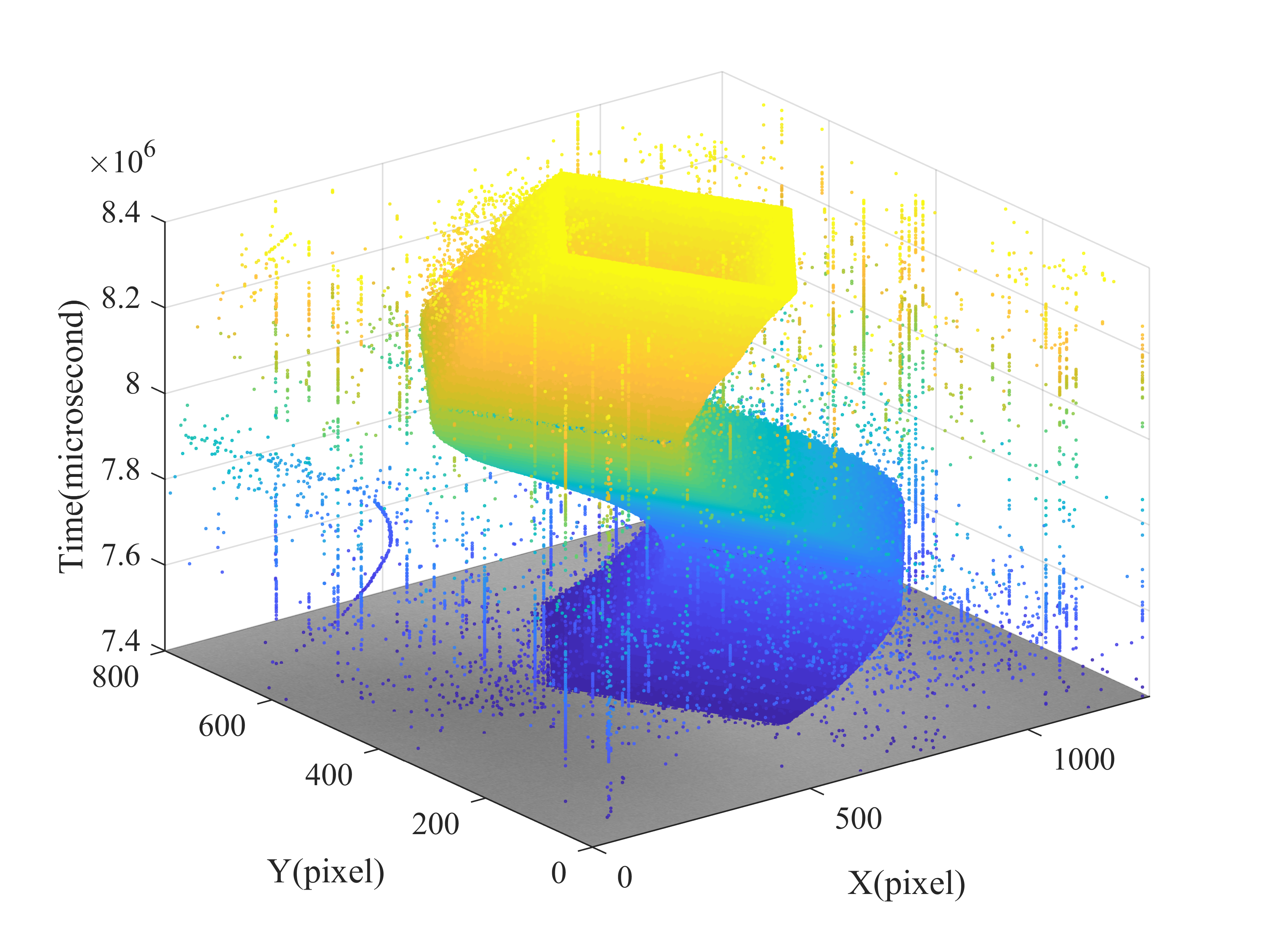} 
		\label{img:celex_GF}
	}
	\subfigure[eFAST$^{*}$]{ 
		\includegraphics[width=1.55in]{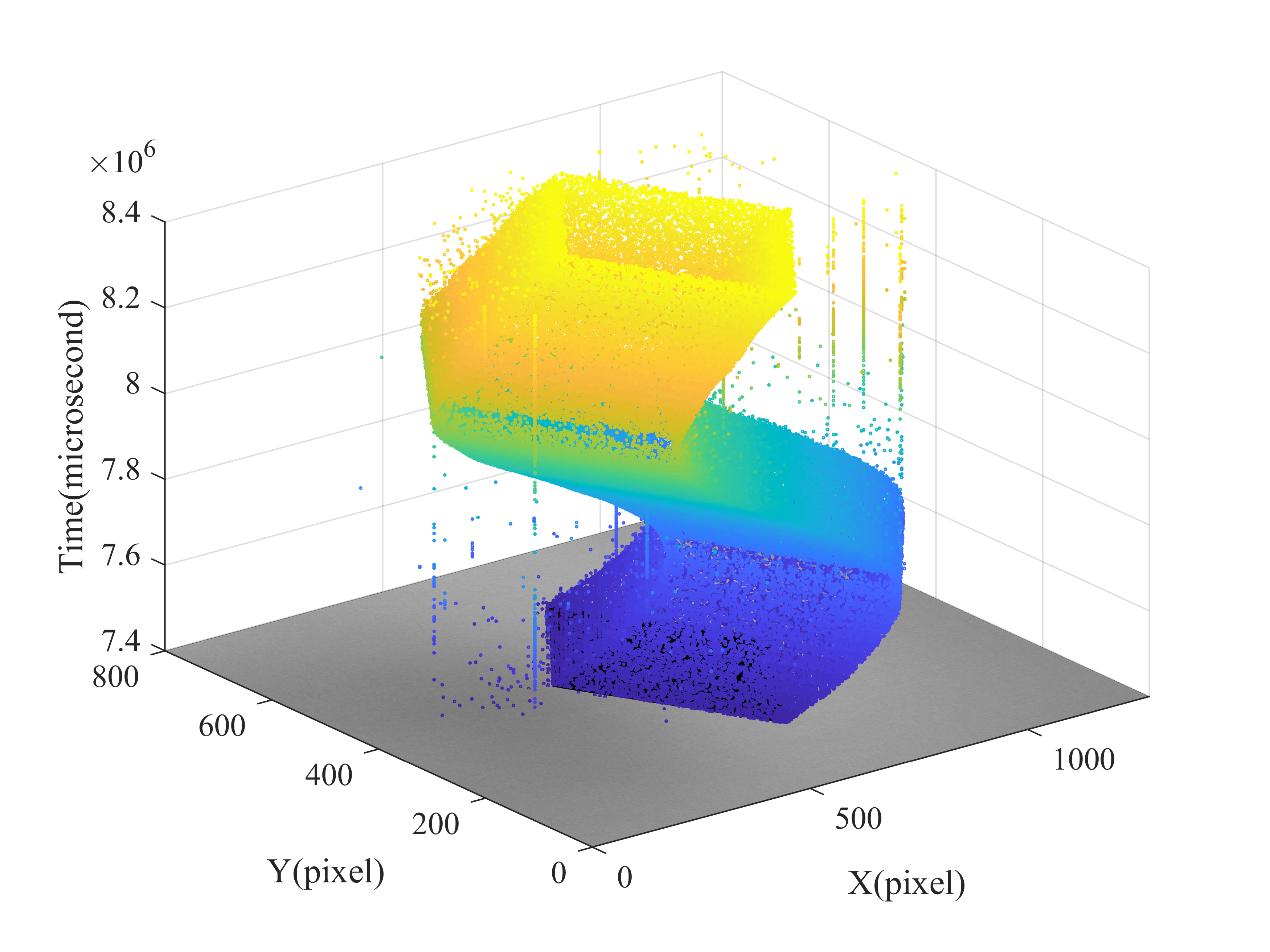} 
		\label{img:celex_efast}
	}
	\subfigure[eSUSAN$^{*}$]{ 
		\includegraphics[width=1.55in]{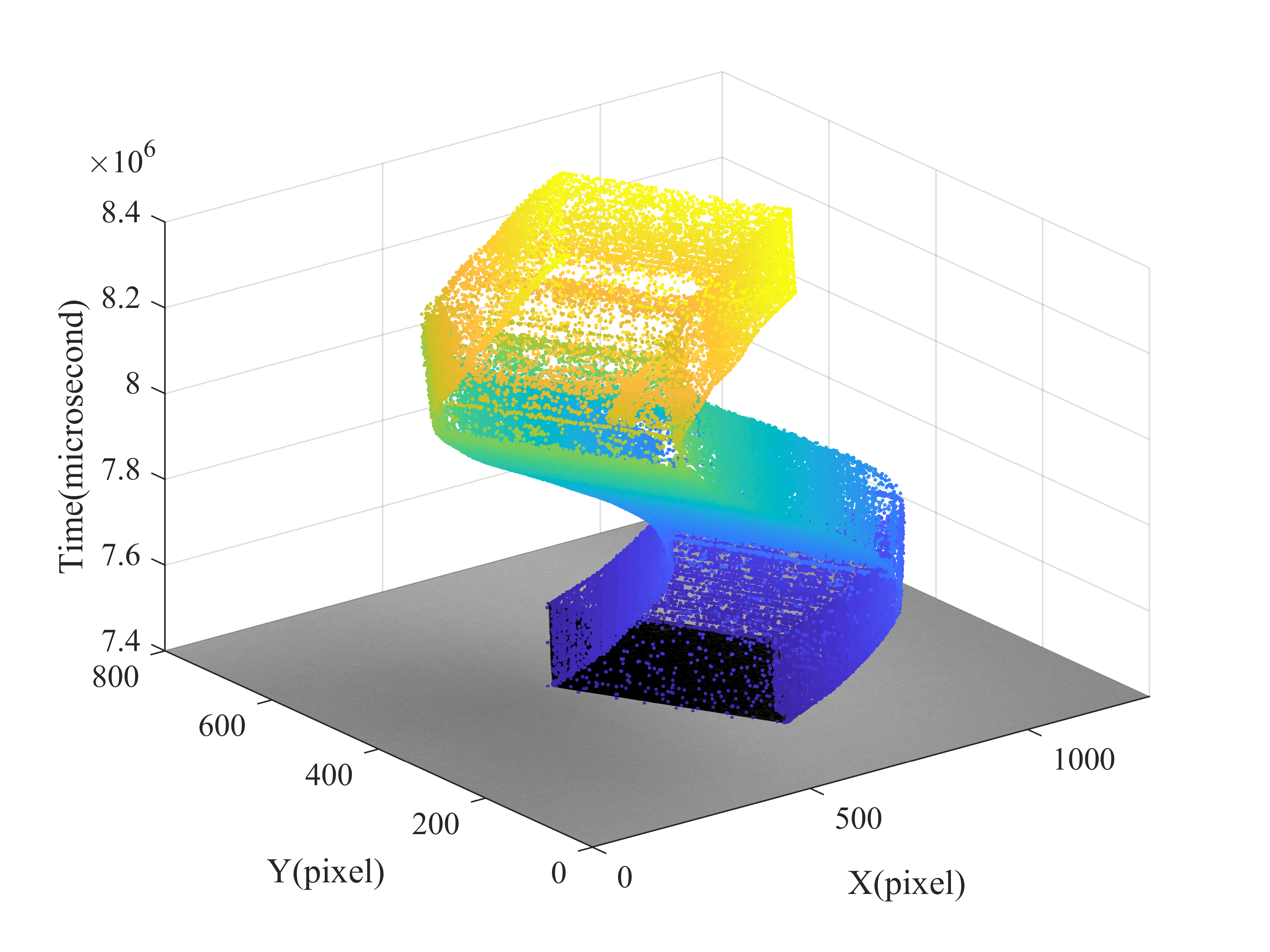} 
		\label{img:celex_esusan}
	}
	\subfigure[G-eHarris$^{*}$]{ 
		\includegraphics[width=1.55in]{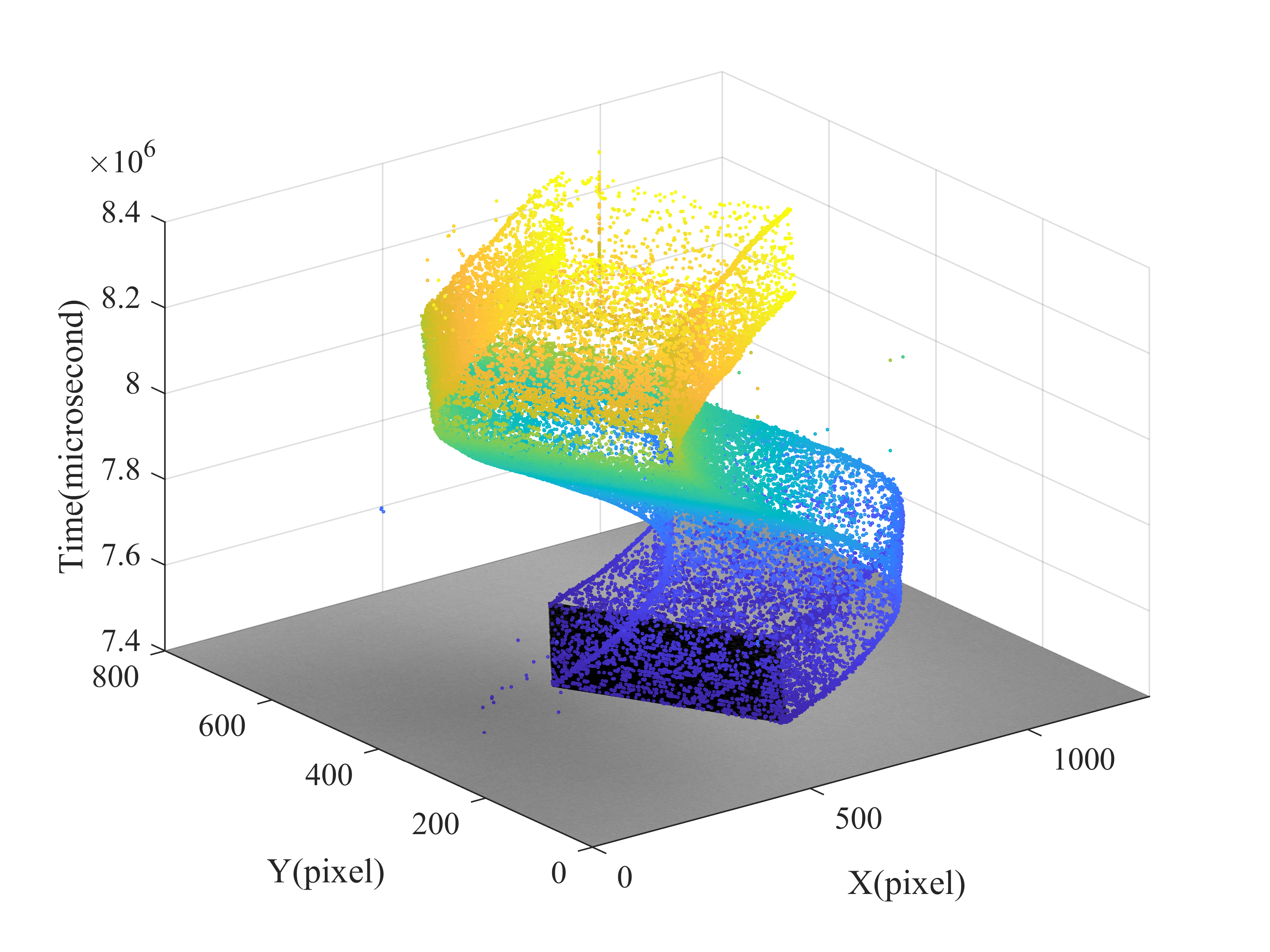} 
		\label{img:celex_geharris}
	}
	\subfigure[AED-eHarris$^{*}$]{ 
		\includegraphics[width=1.55in]{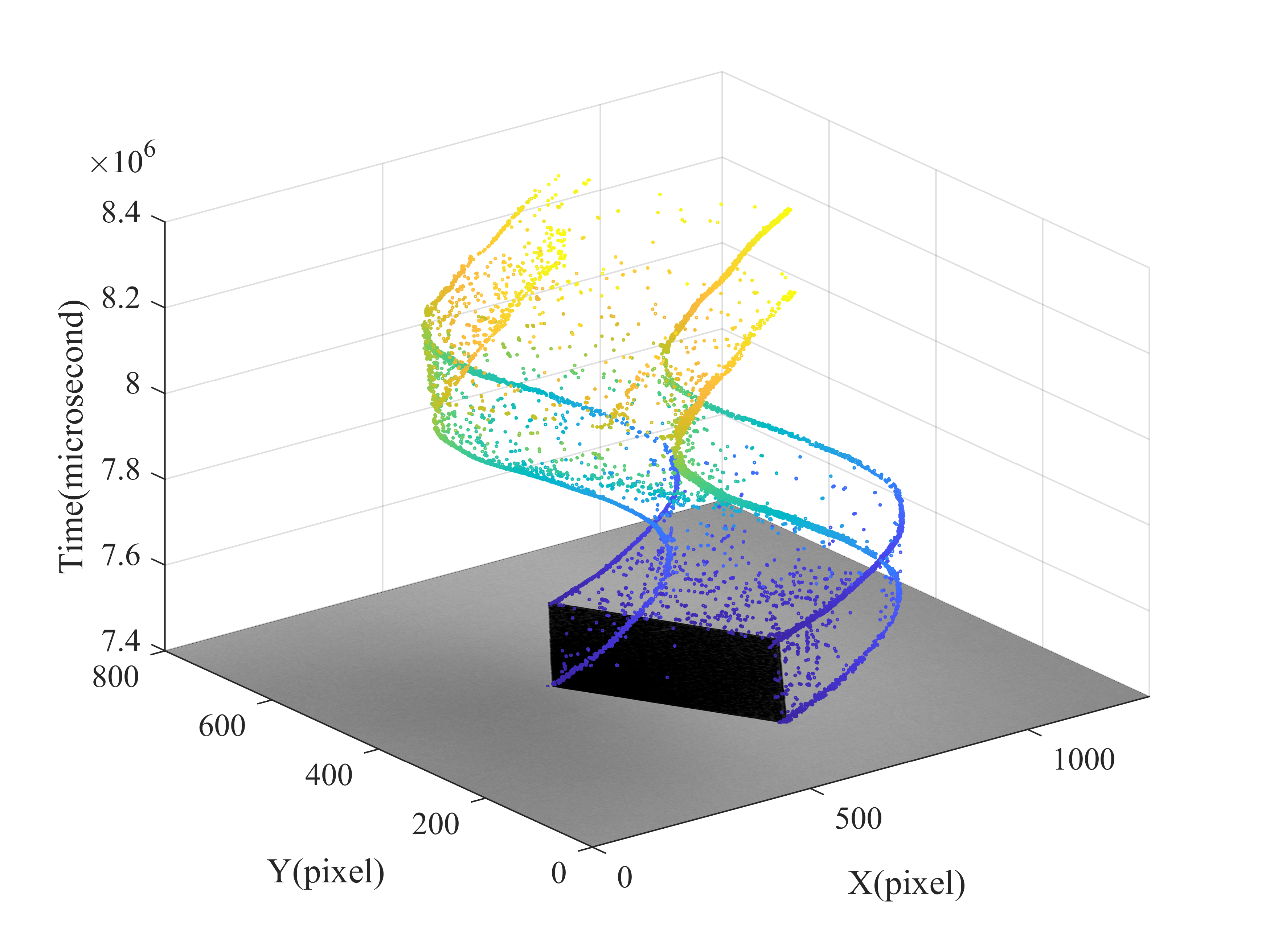} 
		\label{img:celex_aedeharris}
	}
	\captionsetup{font={small}}
	\caption{
		Results of various detectors on CeleX-V dataset. The event stream is formed by a thin square surface that moves quickly under complex lighting conditions. (a) Data seems disordered without obvious temporal and spatial characteristics. (b) Results of GF-filter. Results of (c) eFAST* and (d) eSUSAN*. eSUSAN achieves a higher performance under intense noise. Results of (e) G-eHarris* and (f) EXP-eHarris*. EXP-eHarris* is the only detector that can show corner flow.}
	\label{img:celex_corners}
\end{figure}

We implemented the proposed algorithms using C++ and compared them with existing event-based corner detectors, whose codes were provided by their authors. We also ported various algorithms to MATLAB and evaluated them on the CeleX-V dataset. We performed the evaluation on a computer equipped with a 1.80 GHz Intel i7-10510U processor and 16 GB memory. For a fair comparison, we applied the same denoising algorithm for preprocessing to all the detectors. For the public dataset, we used eFilter proposed in [28]. We compared eHarris and eFast\footnote{\url{https://github.com/uzh-rpg/rpg\_corner\_events}} with eFilter, which are denoted as eHarris* and eFast*, respectively (* represents adding filtering or denoising, same as the GF on the CeleX-V datasets), and also compared the latest Arc*\footnote{\url{https://github.com/ialzugaray/arc\_star\_ros}} and FA-Harris\footnote{\url{https://github.com/ruoxianglee/fa\_harris}}. The proposed algorithm is denoted as eSUSAN* and the algorithm adding eHarris with the AED-SAE is denoted as SE-Harris.

\subsection{Comparison of eHarris Detectors}
\label{subsec:eharris}

We compared the real-time performance of variants of the event-based Harris detector, including eHarris*, G-eHarris* using global SAE, and AED-eHarris*. As reported in \cite{Ignacio2018}, eHarris cannot suitably detect corners with wide angles, as shown in Figure \ref{img:Geharris_shapes_5_8}, given the shape of the local SAE. To increase the algorithm response to angles larger than 180° and prevent a misleading increase in the event reduction rate, we changed the size of the Sobel kernels of eHarris to 7 × 7 and set the score threshold to 16. The detection results are shown in Figure \ref{img:Geharris_shapes_7_16}. We evaluated the running time of the different Harris algorithms, obtaining the results listed in Table \ref{table:eharris}. The result shows that AED-eHarris* is approximately 8 times faster than eHarris* and 2.5 times faster than G-eHarris. It is worth noting that EXP-eHarris can handle medium-textured scenes, such as those in subset dynamic.

\begin{table}[htbp]
	
	\centering
	\captionsetup{font={small}}
	\caption{Running time of eHarris variants. The proposed algorithm is the fastest and the only gradient-based one that can handle medium-texture scenes in real time.}
	\label{table:eharris}
	\begin{tabular}{c|ccc}
		\toprule[1pt]
		\multirow{2}{*}{\textbf{Algorithm}}                                  & \multicolumn{3}{c}{\textbf{\begin{tabular}[c]{@{}c@{}}Total time {[}s{]}\end{tabular}}} \\ \cline{2-4} 
		& shapes                      & dynamic                       & poster                      \\ \hline
		\begin{tabular}[c]{@{}c@{}}eHarris*\cite{Vasco2016}\end{tabular}         & 104.54                      & 386.44                        & 863.86                      \\ \hline
		\begin{tabular}[c]{@{}c@{}}G-eHarris*\cite{Li2019}\end{tabular}       & 37.16                       & 117.94              & 299.37             \\ \hline
		\textbf{\begin{tabular}[c]{@{}c@{}}AED-eHarris*\end{tabular}}  & \textbf{13.81}              & \textbf{44.79}                & \textbf{111.76}           \\ 
		\bottomrule[1pt]         
	\end{tabular}
	
\end{table}

\subsection{Evaluation on Public Event Dataset}
\label{subsec:publiceva}

Table \ref{table:reduction} lists the reduction rates of the number of events in each scene, and Table \ref{table:realtime} lists the processing time per event. The proposed eSUSAN* reduces the number of events similarly to Arc* in a simple scene, while its detection speed is approximately twice faster than that of Arc*. In complex scenes such as dynamic and poster, using the default eSUSAN* parameters results in a lower event reduction rate. Analogously, SE-Harris is almost three times faster than FA-Harris because the former uses eSUSAN and AED-eHarris, which are fast algorithms. Meanwhile, SE-Harris maintains the event reduction rate in high-texture scenes despite using eSUSAN. Thus, hybrid methods such as SE-Harris or FA-Harris are reliable for corner detection and can effectively combine the speed of template-based methods and the accuracy of gradient-based methods. The poor performance of eSUSAN for a high-texture scene indicates the need for manual adjustment of the geometric threshold. In Section \ref{subsec:celexeva}, we change the geometric threshold for the CeleX-V dataset to achieve better results in complex scenes.

\begin{table*}[t]
	
	\centering
	\captionsetup{font={small}}
	\caption{Event reduction rate. The two hybrid methods are listed separately on the right-hand side columns. 
	}
	\label{table:reduction}
	\setlength{\tabcolsep}{1mm}{
		\begin{tabular}{c|cccc|cc}
			\toprule[1pt]
			\textbf{Scenes}      & \begin{tabular}[c]{@{}c@{}}eHarris*\cite{Vasco2016}\\ (\%)\end{tabular} & \begin{tabular}[c]{@{}c@{}}eFAST*\cite{mueggler2017fast}\\ (\%)\end{tabular} & \begin{tabular}[c]{@{}c@{}}Arc*\cite{Ignacio2018}\\ (\%)\end{tabular} & \textbf{\begin{tabular}[c]{@{}c@{}}eSUSAN*\\ (\%)\end{tabular}} & \begin{tabular}[c]{@{}c@{}}FA-Harris\cite{Li2019}\\ (\%)\end{tabular} & \textbf{\begin{tabular}[c]{@{}c@{}}SE-Harris\\ (\%)\end{tabular}} \\ \hline
			shape\_translation   & 92.02                                                         & \textbf{92.27}                                              & 88.14                                                     & 90.32                                                           & 93.48                                                    & \textbf{95.79}                                                    \\
			shapes\_rotation     & \textbf{91.88}                                                & 91.83                                                       & 87.69                                                     & 89.59                                                           & 93.21                                                    & \textbf{95.59}                                                    \\
			shapes\_6dof         & 91.73                                                         & \textbf{92.48}                                              & 88.34                                                     & 89.75                                                           & 93.57                                                    & \textbf{95.75}                                                    \\ \hline
			dynamic\_translation & 94.20                                                         & \textbf{97.53}                                              & 92.71                                                     & 84.09                                                           & 97.54                                                    & \textbf{97.94}                                                    \\
			dynamic\_rotation    & 92.59                                                         & \textbf{96.73}                                              & 91.74                                                     & 82.70                                                           & 96.71                                                    & \textbf{97.48}                                                    \\
			dynamic\_6dof        & 93.60                                                         & \textbf{96.99}                                              & 92.22                                                     & 82.99                                                           & 97.11                                                    & \textbf{97.61}                                                    \\ \hline
			poster\_translation  & 90.03                                                         & \textbf{96.38}                                              & 90.21                                                     & 81.88                                                           & 96.20                                                    & \textbf{97.08}                                                    \\
			poster\_rotation     & 88.75                                                         & \textbf{95.51}                                              & 89.92                                                     & 80.34                                                           & 95.50                                                    & \textbf{97.02}                                                    \\
			poster\_6dof         & 89.15                                                         & \textbf{95.90}                                              & 90.21                                                     & 80.89                                                           & 95.79                                                    & \textbf{97.24}         \\
			\bottomrule[1pt]                                          
	\end{tabular}}
\end{table*}


\begin{table*}[]
	\centering
	\captionsetup{font={small}}
	\caption{Real-time performance. The maximum scene speed was present on subset \emph{shape}. 
	}
	\label{table:realtime}
	\setlength{\tabcolsep}{1mm}{
		\begin{tabular}{c|cccc|cc}
			
			\toprule[1pt]
			\textbf{\begin{tabular}[c]{@{}c@{}}Real-time \\ performance\end{tabular}}      & \begin{tabular}[c]{@{}c@{}}eHarris*\cite{Vasco2016}\end{tabular} & \begin{tabular}[c]{@{}c@{}}eFAST*\cite{mueggler2017fast}\end{tabular} & \begin{tabular}[c]{@{}c@{}}Arc*\cite{Ignacio2018}\end{tabular} & \textbf{\begin{tabular}[c]{@{}c@{}}eSUSAN*\end{tabular}} & \begin{tabular}[c]{@{}c@{}}FA-Harris\cite{Li2019}\end{tabular} & \textbf{\begin{tabular}[c]{@{}c@{}}SE-Harris\end{tabular}} \\ \hline
			\textbf{\begin{tabular}[c]{@{}c@{}}Time per event\\($\mu$s/event)\end{tabular}}   & 5.34                                                          & 0.43                                                        & 0.14                                                      & \textbf{0.06}                                                   & 0.66                                                     & \textbf{0.23}                                                     \\ 
			\textbf{\begin{tabular}[c]{@{}c@{}}Max. Event Rate\\ (Mev/s)\end{tabular}} & 0.20                                                          & 2.44                                                        & 8.93                                                      & \textbf{17.86}                                                  & 1.61                                                     & \textbf{4.76}   \\
			\bottomrule[1pt]                                                 
	\end{tabular}}
\end{table*}

\begin{figure} [t]
	\centering 
	   \subfigure{ 
	     \includegraphics[width=1.55in]{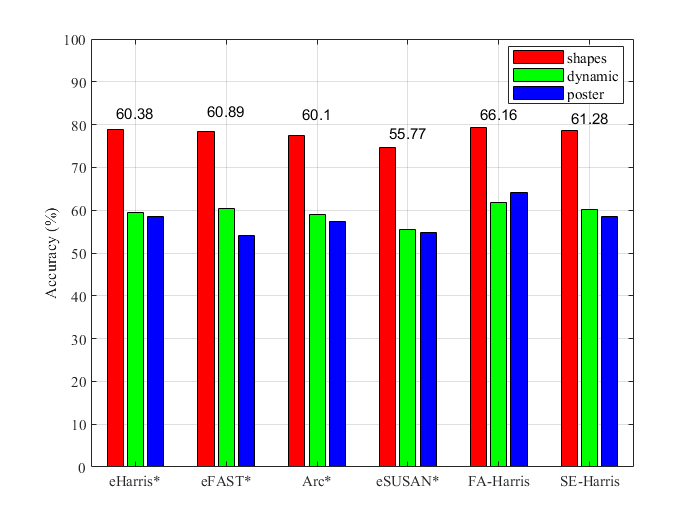} 
	     \label{img:acc}
	   }
	\subfigure{ 
		\includegraphics[width=1.55in]{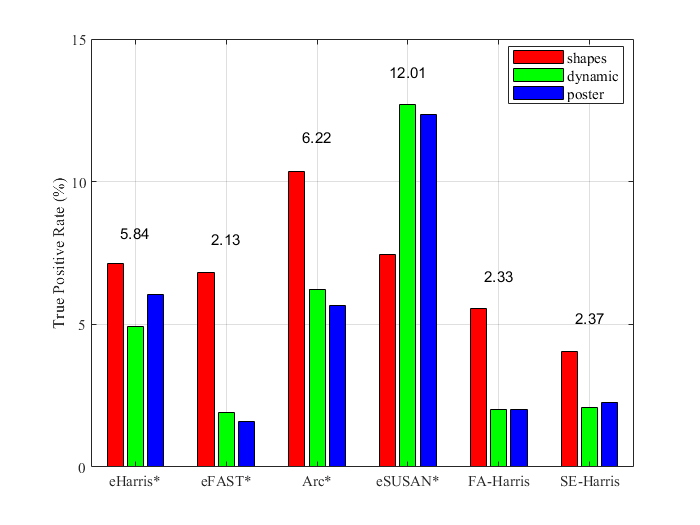} 
		 \label{img:tpr}
	} 
	\captionsetup{font={small}}
	\caption{Accuracy and TPR. We compared the accuracy and TPR of six methods, and each method was tested in three scenes. The results are presented as bar graphs. The numbers indicate the overall performance.
	}
	\label{img:tpracc}
\end{figure}

We evaluated the accuracy and TPR of various detectors, obtaining the results shown in Figure \ref{img:tpracc}. From the detectors, the gradient-based method achieves the highest accuracy. Although eSUSAN* has a similar accuracy to other algorithms in simple scenes, it performs poorly in complex scenes. The accuracy of SE-Harris is only below that of FA-Harris, but its detection is faster. The accuracy and event reduction rate may be closely related, as detectors with a high event reduction rate often achieve higher accuracies.

The TPR is the proportion of corners in the inner cylinder to the total events in the inner cylinder. Thus, a higher TPR indicates that more corners are correctly detected. eSUSAN* achieves an outstanding TPR followed by Arc*. Comparing the two hybrid methods, the overall TPR of SE-Harris is higher because eSUSAN* retains more true corners. On the other hand, the TPR of the hybrid method is low, as also mentioned in \cite{Ylmaz2021}.
In general, SE-Harris achieves an outstanding performance in all aspects. It provides a fast detection with high accuracy and TPR while reducing the number of events.

\subsection{Evaluation on CeleX-V Dataset}
\label{subsec:celexeva}

We also evaluated the performance of various algorithms on the CeleX-V dataset, obtaining the results listed in Table \ref{table:celexresult}. Figure \ref{img:celex_corners} shows a extracted 1 s corner event stream. As the CeleX-V event camera provides no-polarity information, generates a great many noise events, and constructs a single layer SAE, we reduced the scope of corner judgment to achieve the best results by adjusting the default geometric threshold, $g$, of eSUSAN*. The proposed eSUSAN* algorithm filters out more than 98\% of the data in the shortest time, being much faster than eHarris* and eFAST*, and can provide corner detection results close to those algorithms based on the gradient (Figure \ref{img:celex_esusan}). AED-Harris* gets the most intuitive corner flow information, and it is almost immune to noise. SE-Harris can further refine the corners at a faster speed (Figure \ref{img:SE_Harris}). Meanwhile, the detection validity of eSUSAN* is close to that of eHarris* and eFAST*, but eFAST* is severely affected by noise. This is possibly due to the inability to use eFilter on the CeleX-V dataset, leading to a disordered distribution of timestamps on the circles of eFAST*. Moreover, each algorithm has a similar trajectory lifetime. Overall, the proposed detectors provide suitable corner detection for data acquired from a high-resolution event camera.

\begin{table*}[t]
	\centering
	\captionsetup{font={small}}
	\caption{Evaluation results on CeleX-V dataset. The best results are highlighted in boldface.
	}
	\label{table:celexresult}
	\setlength{\tabcolsep}{2mm}{
		\begin{tabular}{c|c|cccc}
			\toprule[1pt]
			\textbf{}                    & GF\cite{Guo2020}    & G-eHarris*\cite{Li2019} & \textbf{AED-eHarris*} & eFAST*\cite{mueggler2017fast}          & \textbf{eSUSAN*} \\ \hline
			\textbf{\begin{tabular}[c]{@{}c@{}}Reduction(\%)\end{tabular}}       & 34.519 & 96.866     & \textbf{98.602}       & 93.701              & 95.630          \\
			\textbf{\begin{tabular}[c]{@{}c@{}}Max. Event Rate (Mev/s)\end{tabular}}       & 19.630  & 0.051    & 0.086            & 0.026             & \textbf{3.235}   \\
			\textbf{\begin{tabular}[c]{@{}c@{}}Average  Lifetime(ms)\end{tabular}} & -     & 11.587     & 12.958                & \textbf{13.540}     & 13.518          \\
			\textbf{\begin{tabular}[c]{@{}c@{}}Valid(\%)\end{tabular}}           & -     & 39.991    & \textbf{64.660}       & 30.927             & 41.826           \\
			\bottomrule[1pt]     
	\end{tabular}
	}
\end{table*}

\section{CONCLUSIONS}
\label{sec:conclusions}

We improve adaptive time threshold $TGF$ using global information to determine the number of similar pixels around an event and implement the SUSAN algorithm to handle event camera data. We demonstrate the effectiveness and real-time performance of the proposed eSUSAN, which is faster than existing corner detection algorithms that can be applied to high-resolution event cameras. For simple objects, the overall performance of eSUSAN is excellent, suggesting its suitability for applications such as visual servoing. In addition, the geometric threshold of eSUSAN can be adjusted to obtain the desired corner type. We also propose an improved normalization method, AED-SAE, for quickly normalizing local SAE while considering changes in scene speed. Based on AED-SAE, SE-Harris achieves real-time performance for corner detection unlike the existing gradient-based method. In future work, we will further investigate denoising of data from high-resolution event cameras and improve the adaptive time threshold and AED-SAE, aiming to extend their use to other vision systems.



\section*{ACKNOWLEDGMENT}
\label{sec:acknowledgment}
Many thanks to Elias Mueggler, Ignacio Alzugaray and Ruoxiang Li, who provided their source code for event-based corner detection, and also very grateful to related researchers for their inspiration to us.

\bibliographystyle{IEEEtran}
\bibliography{IEEEabrv,references}
\end{document}